\documentclass[sigconf]{acmart}
\AtBeginDocument{%
  }
\usepackage{bm}
\usepackage{algorithm}
\usepackage{algorithmic}
\AtBeginDocument{%
  }

\usepackage{color,xcolor}
\usepackage{amsmath}
\usepackage{marvosym}
\setcopyright{acmlicensed}
\copyrightyear{2025}
\acmYear{2025}
\acmDOI{10.1145/3711896.3737135}
\acmConference[KDD '25]{Proceedings of the 31st ACM SIGKDD Conference on Knowledge Discovery and Data Mining V.2}{August 3--7, 2025}{Toronto, ON, Canada}
\acmBooktitle{Proceedings of the 31st ACM SIGKDD Conference on Knowledge Discovery and Data Mining V.2 (KDD '25), August 3--7, 2025, Toronto, ON, Canada}
\acmISBN{979-8-4007-1454-2/2025/08}
\begin{document}

\title{SSD-TS: Exploring the Potential of Linear State Space Models for Diffusion Models in Time Series Imputation}

\author{Hongfan Gao}
\affiliation{%
  \institution{East China Normal University\textsuperscript{1}}
  \city{Shanghai}
  \country{China}
}
\email{hf.gao@stu.ecnu.edu.cn}

\author{Wangmeng Shen}
\affiliation{
    \institution{East China Normal University\textsuperscript{1}}
    \city{Shanghai}
    \country{China}
}

\email{wmshen@stu.ecnu.edu.cn}

\author{Xiangfei Qiu}
\affiliation{
    \institution{East China Normal University\textsuperscript{1}}
    \city{Shanghai}
    \country{China}
}
\email{xfqiu@stu.ecnu.edu.cn}

\author{Ronghui Xu}
\affiliation{
    \institution{East China Normal University\textsuperscript{1}}
    \city{Shanghai}
    \country{China}
}
\email{rhxu@stu.ecnu.edu.cn}

\author{Bin Yang}
\affiliation{
    \institution{East China Normal University\textsuperscript{1}}
    \city{Shanghai}
    \country{China}
}
\email{byang@dase.ecnu.edu.cn}

\author{Jilin Hu \Letter}
\affiliation{
    \institution{East China Normal University\textsuperscript{1,}\textsuperscript{2}}
    \city{Shanghai}
    \country{China}
}
\email{jlhu@dase.ecnu.edu.cn}

\renewcommand{\shortauthors}{Hongfan Gao et al.}

\begin{abstract}
Probabilistic time series imputation has been widely applied in real-world scenarios due to its ability for uncertainty estimation and denoising diffusion probabilistic models~(DDPMs) have achieved great success in probabilistic time series imputation tasks with its power to model complex distributions.  However, current DDPM-based probabilistic time series imputation methodologies are confronted with two types of challenges: 1)\textit{The backbone modules of the denoising parts are not capable of achieving sequence modeling with low time complexity.} 2)~\textit{The architecture of denoising modules can not handle the dependencies in the time series data effectively.} To address the first challenge, we explore the potential of state space model, namely Mamba, as the backbone denoising module for DDPMs. To tackle the second challenge, we carefully devise several SSM-based blocks for time series data modeling. Experimental results demonstrate that our approach can achieve state-of-the-art time series imputation results on multiple real-world datasets. Our datasets and code are available at \href{https://github.com/decisionintelligence/SSD-TS/}{https://github.com/decisionintelligence/SSD-TS/}
\footnotetext[1]{School of Data Science and Engineering.}
\footnotetext[2]{Engineering Research Center of Blockchain Data Management, Ministry of Education.}
\end{abstract}

  \ccsdesc[500]{Mathematics of computing~Time series analysis}
  \ccsdesc[300]{Computing methodologies~Neural networks}

\keywords{Diffusion Models, State Space Models, Time Series Imputation}

\newcommand\kddavailabilityurl{https://doi.org/10.5281/zenodo.15528129}
\maketitle
\ifdefempty{\kddavailabilityurl}{}{
\begingroup\small\noindent\raggedright\textbf{KDD Availability Link:}\\
The source code of this paper has been made publicly available at \url{\kddavailabilityurl}.
\endgroup
}
\section{Introduction} 
The analysis of time series can model the intrinsic patterns within time-series data, thus providing robust support for decision-making in various fields, such as meteorology~\cite{DBLP:journals/datamine/McGovernRBD11,DBLP:journals/nn/KarevanS20,DBLP:conf/icde/00010GY0HXJ24}, financial analysis~\cite{DBLP:conf/cikm/XiangCSZL22,DBLP:conf/icdm/OwusuTPBW23,DBLP:conf/ijcai/0001CW0H20,li2024TSMF-Bench}, healthcare~\cite{DBLP:journals/tmis/MoridSD23,DBLP:conf/ml4h/PoyrazM23,DBLP:journals/corr/abs-2503-04252}, power systems~\cite{DBLP:conf/icassp/TzelepiNT23,DBLP:conf/aaai/ZhouZPZLXZ21,DBLP:conf/aaai/ZhangSCLGY25,DBLP:journals/pvldb/ZhaoGCHZY23,DBLP:journals/pacmmod/Wu0ZG0J23} and traffic~\cite{DBLP:journals/sigmod/GuoJ014,10.1145/3690624.3709209,xu2023spatial,DBLP:conf/iclr/LiuSGY25,Air-DualODE}. To enhance the reliability of analytical outcomes, it is critical to ensure the integrity of time series. However, due to various reasons such as device failures, human errors, and privacy protection, time series data can easily be incomplete with missing observations at different timestamps. 

Time series imputation methods aim to estimate the values of missing points based on the observed points in incomplete time series, thereby restoring the integrity of the time series while preserving its original statistical properties. According to the ability to provide uncertainty of estimations, time series imputation methods can be categorized into the following two perspectives: 1)~\textit{Deterministic}~\cite{DBLP:conf/nips/CaoWLZLL18,DBLP:conf/iclr/CiniMA22,DBLP:journals/eswa/DuCL23}, and 2)~\textit{ Probabilistic}~\cite{DBLP:conf/icml/ChenDFLYZRZSN23,DBLP:conf/icml/KimKYLL023,DBLP:conf/nips/LuoCZXY18} imputation methods. \textit{Probabilistic} time series imputation is particularly important in dealing with complex and uncertain data environments, as it provides a quantification of uncertainty for the imputations. 
The key to probabilistic imputation lies in modeling the posterior distribution. Existing probabilistic time series imputation methods include \textit{Gaussian Process and Variational Autoencoder}-based methods~\cite{DBLP:conf/aistats/FortuinBRM20}, \textit{Normalization Flow}-based methods~\cite{DBLP:conf/iclr/RasulSSBV21}, and \textit{Diffusion}-based methods~\cite{DBLP:conf/nips/TashiroSSE21}. Among these, the \textit{Diffusion}-based method has emerged as the optimal choice for probabilistic time series due to their accuracy in posterior modeling and adaptability to different scenarios and various types of time series data. 

\begin{table}[tbp]
\caption{Comparison of our method and existing methods in modeling dependencies and time complexity. }
\label{tab:backbonecmp}
\resizebox{\linewidth}{!}{
\begin{tabular}{l|cccc}
\hline
Backbone Model            & CNN              & Transformer        & SSM              & SSD-TS~(Ours)    \\ \hline
Global Dependency         & Local            & Global             & Partial          & Global           \\
Time Complexity           & $\mathcal{O}(L)$ & $\mathcal{O}(L^2)$ & $\mathcal{O}(L)$ & $\mathcal{O}(L)$ \\
Channel Dependecny        & Independent      & Independent        & Independent      & Dependent        \\
Inter-sequence Dependency & Unidirectional   & Unidirectional     & Unidirectional   & Bidirectional    \\ \hline
\end{tabular}}
\end{table}

When selecting a denoising backbone in the diffusion model, the following two key factors need to be considered: 1)~\textbf{Model compatibility}, and 2)~\textbf{Time complexity}. Model compatibility involves two key aspects: 1)~the backbone of the model should be capable of handling input data effectively. 2)~the backbone of the model should align with the model's intended objective~(\textit{i.e.}, in diffusion models, the backbone must be capable of modeling noise in the diffusion process).
Specifically, the missing observations in time series have correlations with their neighbors on both sides, \textit{so it is crucial to design a model by considering information from neighbors of both sides}. Moreover, \textit{it is also essential to accurately capture the properties of time series}, such as global dependencies and channel correlations. 
Three mainstream denoising backbones are widely used in diffusion models for time series imputation: 1)~\textit{Convolutional Neural Networks~(CNNs)}-based, 2)~\textit{Transformer}-based and 3)~\textit{State-Space Model~(SSM)}-based backbones. Given a time series with a length of $L$, \textit{the CNNs-based backbone} can capture partial information from the neighbors within the receptive fields and has $\mathcal{O}(L)$ time complexity. \textit{The transformer-based backbone} can model temporal dependencies across the entire time series but is with quadratic time complexity $\mathcal{O}(L^2)$. \textit{The SSM backbone} has a linear time complexity, $\mathcal{O}(L)$, but it falls short in capturing the information from one side of the neighbor. Moreover, all these backbones fail to capture the channel dependencies in time series. The comparison results of existing backbones and our method in terms of various dependencies and time complexity are presented in Table.\ref{tab:backbonecmp}.

Recently, Mamba~\cite{DBLP:conf/icml/DaoG24}, a linear sequence modeling approach based on state-space models has demonstrated superior performance in various time series tasks, including time series forecasting, imputation and time series foundation models~\cite{DBLP:journals/ijon/WangKFWYZWZ25, DBLP:journals/corr/abs-2411-02941}. Compared to Transformer-based models, Mamba-based models have shown stronger sequence modeling capabilities and better performance in time series tasks. However, there has not been research investigating the Mamba model with diffusion models for time series applications. Whether the Mamba model can serve as the backbone of a diffusion model and achieve competitive performance remains an open question.

In this paper, we investigate how to apply Mamba as backbones for time series diffusion models. As shown in Table.\ref{tab:backbonecmp}, existing backbones exhibit issues in sequence dependency modeling, time complexity, and inter-channel dependency modeling. To address time complexity, we adopt Mamba of linear complexity, as the foundation for our base module. To tackle intra-sequence and bidirectional dependency modeling, we propose the Bidirectional Attention Mamba~(BAM) block, a bidirectional module with temporal attention based on the linear state-space model Mamba, achieving effective and efficient intra-channel dependency modeling. Furthermore, for inter-channel dependencies, we analyze the inductive biases inherent in inter-channel relationships and compare how these biases are captured by the Mamba-based channel modeling module versus CNN- and transformer-based approaches. 
Our findings indicate that the Mamba-based module provides a more accurate and comprehensive representation, making it a superior choice. Based on this analysis, we design the Channel Mamba Block~(CMB) to effectively model inter-channel dependencies.

Our contributions are summarized as follows:
\begin{enumerate}
    \item We propose SSD-TS, a \textbf{S}tate \textbf{S}pace \textbf{D}iffusion model for \textbf{T}ime \textbf{S}eries imputation. It integrates mamba-based blocks as diffusion backbones and equips the model with the capability of probabilistic time series imputation with linear time and space complexity.
    \item We give a brief analysis of the characteristics and inductive biases of both intra-channel and inter-channel dependencies, thereby empirically elucidating the superiority of employing mamba modules as a backbone for diffusion models. Based on these analyses, we propose bidirectional attention mamba~(BAM) block and channel mamba block~(CMB) to achieve effective modeling of these dependencies.
    \item We conduct experiments on multiple real-world datasets for time series imputation task. Our approach achieves state-of-the-art performance across several datasets, different missing scenarios and missing ratios, which demonstrate the effectiveness of our proposed method.
\end{enumerate}
The rest of our paper is organized as follows: In Section \ref{sec:pre}, we present preliminaries of state space models and diffusion models as well as the problem formulation. In Section \ref{sec:method}, we give a brief introduction about the diffusion models and the architecture details about our model. In Section \ref{sec:exp}, we present and analyze the experiment results. In Section \ref{sec:rel}, we summarize the related work and in Section \ref{sec:con}, we conclude our paper.

\section{Preliminaries}
\label{sec:pre}
\subsection{State Space Models}
State Space Models (SSMs) are an emerging approach to model sequential data,
which is implemented by finding out
state representations to model the
relationship between input and
output sequences. An SSM receives a one-dimensional sequence
$X\in\mathbb{R}^{L}$ as the input and outputs a corresponding
sequence $Y\in\mathbb{R}^{M}$. Under continuous settings, the SSMs are defined 
according to Eq.\ref{eq:ssm1}:
\begin{equation}
  \begin{cases}\dot{h}(t)&=\bm{A}h(t)+\bm{B}x(t)\\y(t)&=\bm{C}h(t)+\bm{D}x(t),\end{cases}
  \label{eq:ssm1}
\end{equation}
where $x(t)\in\mathbb{R}^{L}$, $y(t)\in\mathbb{R}^{M}$, $h(t)$, and $\dot{h}(t) \in \mathbb{R}^{N}$ stands for the input, output, hidden state, and derivative of hidden state at timestamp $t$, respectively; $\bm{A}\in\mathbb{R}^{N\times N},\bm{B}\in\mathbb{R}^{N\times L},\bm{C}\in\mathbb{R}^{M\times N}$ and $\bm{D}\in\mathbb{R}^{M\times L}$
are learnable model parameters. 

In real-world applications, the input sequences are discrete samplings of continuous sequences. According to ~\cite{DBLP:conf/iclr/GuGR22}, under discrete settings, by applying the zero-order hold technique to Eq.\ref{eq:ssm1},
it can be reformulated as follows. 
\begin{equation}
    \begin{cases}h_k = \bm{\bar{A}}h_{k-1}+\bm{\bar{B}}x_{k}\\y_k = \bm{C}h_{k}\end{cases},
    \label{eq:disssm}
\end{equation}
where $\bm{\bar{A}}=\exp(\Delta \bm{A}), \bm{\bar{B}}=(\Delta \bm{A})^{-1}(\exp(\Delta \bm{A})-\bm{I})\cdot(\Delta \bm{B})$ 
and $\Delta$ is the learnable step size in discrete sampling. We can see from Eq.\ref{eq:disssm} that the hidden state is updated according to the input $x(t)$ and last hidden state $h(t-1)$ while the output is generated by the hidden state $h(t)$ and the input $x(t)$ and in~\cite{DBLP:conf/nips/GuDERR20}, where it introduces High-order Polynomial Projection Operator (Hippo) to achieve longer sequence modeling. 

However, it is worth noticing that $\bm{A},\bm{B},\bm{C},\bm{D}$ in Eq.\ref{eq:ssm1} and Eq.\ref{eq:disssm} are time-invariant parameters, \textit{i.e.}, they are data-independent parameters and do not change over time. Therefore the model is not capable of assigning different weights at different positions in the input sequence while receiving new inputs.
To address this issue,~\cite{DBLP:journals/corr/abs-2312-00752} 
proposed Mamba, in which the parameter matrices $\bm{A},\bm{B},\bm{C},\bm{D}$ are input-dependent, thus enhancing the performance of sequence modeling. To tackle the problem of non-parallelization,~\cite{DBLP:journals/corr/abs-2312-00752} also introduced selective scan mechanism for effective computing. For further performance and efficiency improvements,~\cite{DBLP:conf/icml/DaoG24} point out that SSMs can be categorized as a variant of linear attention model.
In this work, we follow the same architecture of parallel Mamba Blocks as~\cite{DBLP:conf/icml/DaoG24} and a RMS-norm~\cite{DBLP:conf/nips/ZhangS19a} module is added after the parallel Mamba block.
The details of the post-normalization Mamba Block (PNM Block) are illustrated in Fig.\ref{fig:modules}a.
\subsection{Diffusion Models}
Let $x_t$ be a sequence of variables for $t = 1,2,\cdots,T$. The diffusion process consists of two processes: 1)~\textbf{The forward process} without learnable parameters, which transforms the data distribution into a standard Gaussian distribution by gradually adding noise to the data. 2)~\textbf{The reverse process} with learnable parameters, which first samples from the standard Gaussian distribution and then progressively denoises the data to approximate the data distribution. The reverse process of diffusion models a parameterized distribution $p_{\theta}$ defined with the following Markov chain to approximate the real data distribution:
\begin{equation}
  p_{\theta}(x_{0:T}) = p(x_T)\prod_{t=1}^{T}p_{\theta}(x_{t-1}\vert x_t),
  \label{eq:rprocess:con}
\end{equation}
where $x_T\sim \mathcal{N}(0,I)$ denotes the latent variable sampled 
from standard Gaussian distribution and 
\begin{equation}
  p_{\theta}(x_{t-1}\vert x_t) = \mathcal{N}(x_{t-1};\mu_{\theta}(x_t,t),\sigma_{\theta}(x_t,t)I),
  \label{eq:rp:con}
\end{equation}
The loss function of DDPM aims at minimizing the difference between the noise $\epsilon$ in the forward process and the parameterized noise $\epsilon_{\theta}$ in the reverse process:
\begin{equation}
  \mathcal{L}_{d} = \mathbb{E}_{x_0,\epsilon}\Vert \epsilon-\epsilon_{\theta}(x_t,t)\Vert,
  \label{eq:loss:con}
\end{equation}
where $t$ stands for the diffusion time embedding and $x_t$ is calculated in the forward process. 
\subsection{Problem Formulation}
\label{subsec:ana}

\begin{definition}[\textbf{Time Series}]
\label{def:incts}
A time series can be defined as a tuple, denoted as $\tilde{X}=(X, M, T)$, where $X \in \mathbb{R}^{K \times L}$ is the observation matrix with $K$ observations at a time, which are ordered along $L$ time intervals chronologically; $M\in \mathbb{R}^{K \times L}$ is an indicator matrix that indicates whether the observation at $(i,j)$ in $X$ is missing or not: if the observation at position $(i, j)$ is missing, i.e., $X_{i, j}=\text{NA}$, then $M_{i,j} = 1 $, otherwise, $ M_{i,j} = 0 $; $T\in \mathbb{R}^{L}$ is the time stamps of the time series. 
\end{definition}


\noindent
\textbf{Problem Statement}~\textbf{(Probabilistic Time Series Imputation).}
Given an incomplete time series $\tilde{X}=(X, M, T)$, the problem of probabilistic time series imputation aims to approximate the real poster distribution $p(\bar{X}\vert\tilde{X})$. Deep learning based probabilistic time series imputation methods learn an imputation function $\mathcal{M}_{\theta}$, such that 
\begin{equation}
    \bar{X} = \mathcal{M}_{\theta}(\tilde{X}),
\end{equation}
where $\bar{X}\in \mathbb{R}^{K \times L}$ is the imputed time series, $\bar{X}_{i,j} = \mu_{i,j} \pm \sigma_{i,j}$ denotes the probabilistic output if $M_{i,j} = 1 $, otherwise $\bar{X}_{i,j} = {X}_{i,j}$. 

\section{Methodology}
\label{sec:method}
\begin{figure}[tbp]
    \centering
    \includegraphics[width=\linewidth]{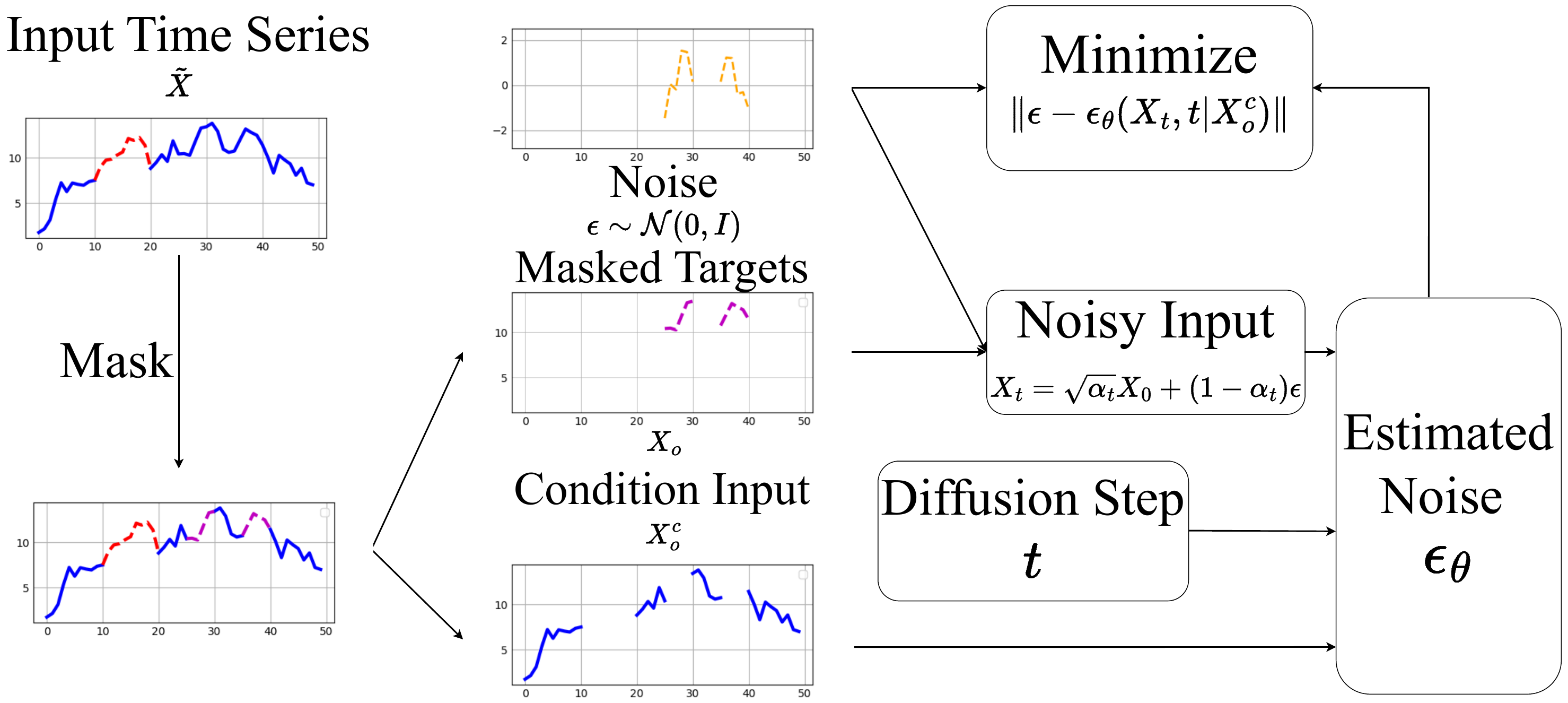}
    \caption{The self-supervised framework and training process of SSD-TS. First, some of observed values are masked following the same missing pattern as the missing values~(in red) to get masked targets~($X_0$, in magenta) and the condition input~($X_o^c$, in blue). The noisy input is obtain from $X_0$ and $\epsilon$~(in orange) sampled from $\mathcal{N}(0,I)$. The objective of the network is to minimize the difference between the parameterized noise $\epsilon_\theta(X_t,t)$ and $\epsilon$. Solid lines in each time series represent observed values, while dashed lines represent missing values.}
    \label{fig:ssl}
\end{figure}


\subsection{Overall Model Architecture}
Fig.\ref{fig:ssl} illustrates the overall self-supervised framework and training process of our model. We first mask part of the observed values according to the pattern of missing values, where the masked values serve as the imputation target $X_0$ during training. The remaining observed values form the conditional input $ X_o^c $ for the noise prediction network $\epsilon_{\theta}$. We then combine $ X_0 $ with noise $\epsilon$ sampled from a standard normal distribution to obtain the noisy input $X_t$.
Both  $X_o^c$, $X_t$, and the diffusion step $t$ are fed into the noise prediction network $\epsilon_\theta$ to get the parameterized noise. 

\begin{figure*}[tbp]
    \centering
    \includegraphics[width=\linewidth]{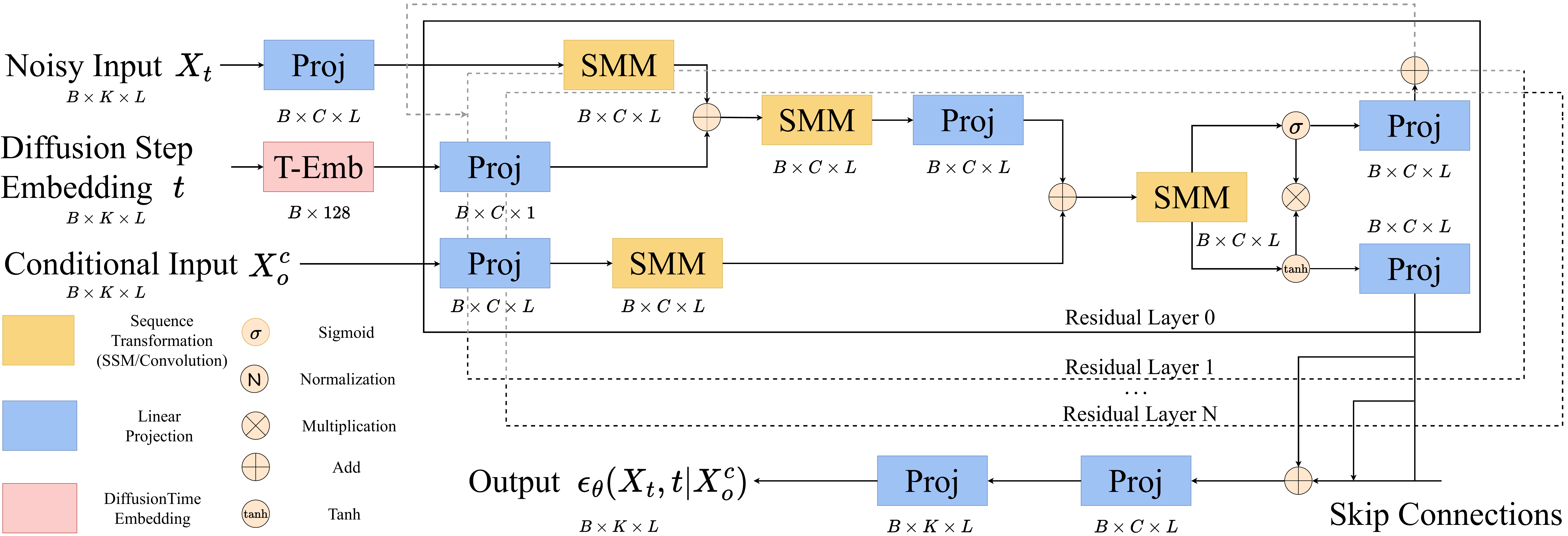}
    \caption{Architecture of noise estimation module $\epsilon_\theta$ in SSD-TS.}
    \label{fig:modelarch}
\end{figure*}

Fig.\ref{fig:modelarch} presents the denoising architecture of our module. The forward process of $\epsilon_\theta$ are as follows: 
For each diffusion step, the input consists of the following parts:
noisy input $X_t$, the condition input $X_o^c$ and the diffusion step $t$.
To begin with, the inputs are embedded to the latent diffusion space. The 
embedding module of noisy inputs and condition inputs share 
a similar model structure, which consists of a linear projection 
module followed by an SMM block in Fig.\ref{fig:modules}b. The SMM block is composed of stacks of Bidirectional Attention Mamba~(BAM) blocks and Channel Mamba Blocks~(CMB), which is introduced in the next part.
Due to the relatively limited information from $t$, the embedding
module of $t$ only consists of linear projection modules. 
After the embedding step, the embedded diffusion step is concatenated
with the input embeddings. The concatenated embeddings are fed in to a
SMM module. Then the output of the SMM module is concatenated with the condition
embeddings. After feeding the final embeddings to another SMM module and
final projection module, we can get the noise predictions $\epsilon_\theta(X_t,t)$.

\subsection{Intra-Channel Dependency Modeling}
We start by analyzing what the denoising module needs to model within a single channel. Since Triformer~\cite{DBLP:conf/ijcai/CirsteaG0KDP22} first proposes the Patching technique and existing works~\cite{DBLP:conf/iclr/NieNSK23} demonstrate the effectiveness of patching in time series. We do not apply the patching technique to the time series as our approach is non-autoregressive, which means that we generate a complete sequence of size $L$ for a total of \(T\) steps ($T < L$). The noise described by the denoising module corresponds to the variations between consecutive steps. This necessitates that the denoising module accurately models the internal dependencies within the sequence in order to precisely predict the noise in the intermediate steps. 

As shown in Table.\ref{tab:backbonecmp}, the main drawback of convolutional neural network (CNN)-based backbones, such as U-Net~\cite{DBLP:conf/miccai/RonnebergerFB15}, lies in their inability to capture global information within the sequence. Additionally, stacking convolutional layers increases time complexity. Transformer-based backbones, on the other hand, suffer from a lack of dynamic weight allocation and exhibit quadratic complexity with respect to sequence length $L$. Moreover, existing SSM-based models rely on unidirectional intra-channel dependency modeling, limiting their ability to capture global dependencies in time series data.

To address these issues, we design an intra-channel dependency modeling module based on a bidirectional linear state space model. Specifically, given an input representation $x$, the BAM module first normalizes the input $x$ using layer normalization.  The normalized features are processed in both forward and backward directions. The forward processing involves a linear convolutional layer followed by a PNM module. The backward sequence shares a similar processing procedure and the input sequence is flipped along the time dimension. After processing in both directions, a temporal attention module is applied to adjust the weights for each time step. Finally, the results from both directions are element-wise summed after aligning them by position, and a residual connection is added to the original input $x$, producing the output of the BAM module. The detailed structure of the BAM module is shown in Fig.\ref{fig:modules}d.

And it is worth noticing the temporal attention module in Fig.\ref{fig:modules}d is not analogous to the attention mechanism used in Transformer. Our temporal attention module takes a sequence as input and applies a convolutional layer followed by a sigmoid activation to produce a set of weights for each position in the sequence. These weights are then multiplied element-wise with the input sequence. The purpose of this module is to adaptively reweight the temporal positions within the sequence, thereby enhancing the modeling of global dependencies and helping the state space model (SSM) better capture long-range temporal relationships.

\subsection{Inter-Channel Dependency Modeling}
Addressing the time series modeling problem, capturing inter-channel dependencies allows for the detection of mutual influences among different channels (variables or features), helping to more accurately reveal the global dynamics and complex interactions of the system, thereby improving predictive performance and generalization ability~\cite{qiu2025duet,wu2024catch}.

\begin{figure}[t]
    \centering
    \includegraphics[width=\linewidth]{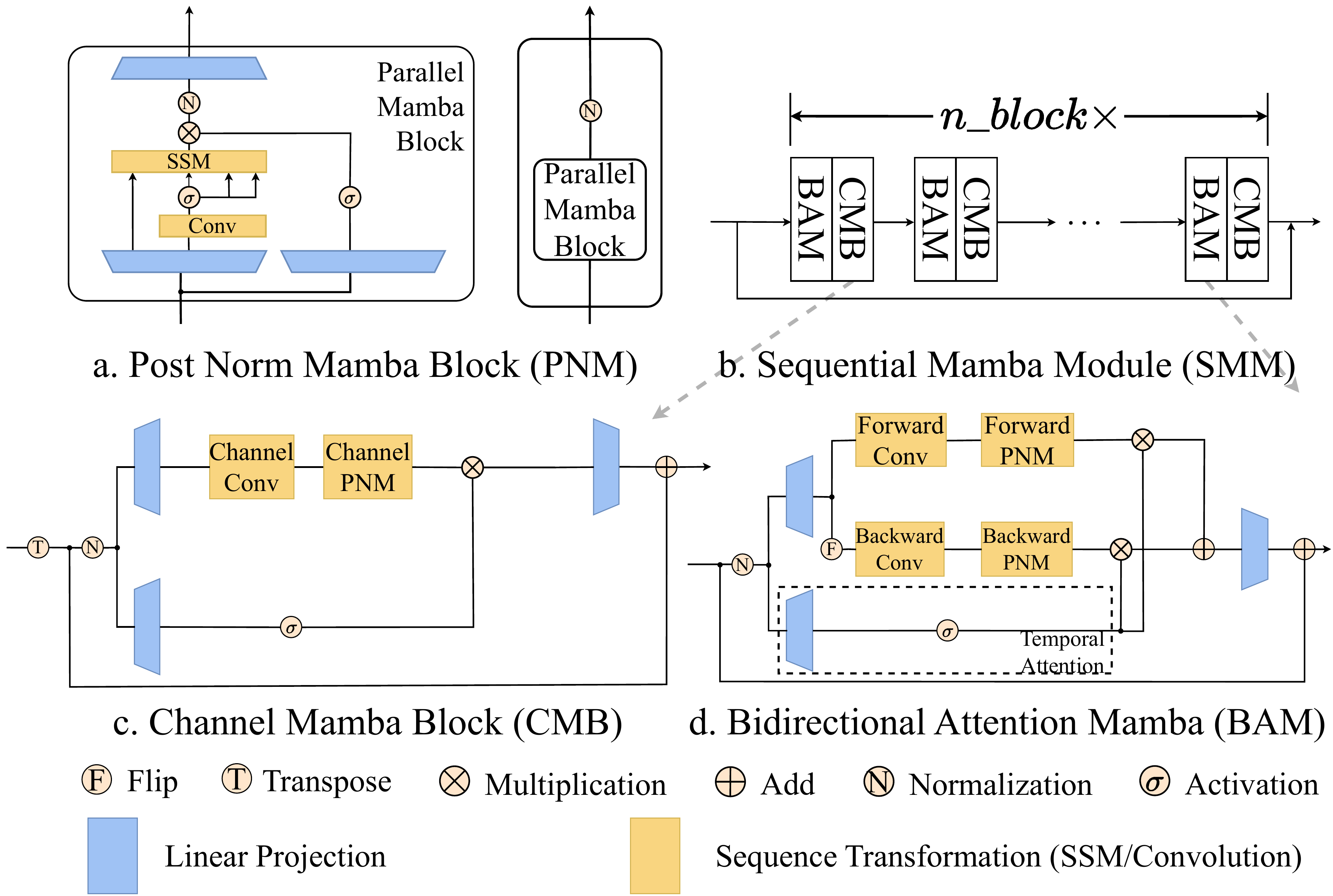}
    \caption{Details of PNM, CMB and BAM block in the noise prediction module. (a)~PNM: backbone module based on Mamba. (b)~SMM: Sequential Mamba module composed of stacks of BAM and CMB. (c)~CMB: unidirectional module for inter-channel dependency modeling. (d)~BAM: bidirectional module with temporal attention for intra-channel, multi-range dependency modeling.}
    \label{fig:modules}
\end{figure}

For more accurate modeling of the channel dependencies, the key is to identify the inductive biases inherent in the dependencies between channels in time series data. Through an analysis of the characteristics of time series data and previous methods~\cite{DBLP:conf/nips/TashiroSSE21,DBLP:conf/iclr/LiuLCCJ23}, we argue that the modules designed for modeling channel relationships should possess the capability to describe the following inductive biases:
\begin{itemize}
    \item \textit{Inter-Channel Dependencies}: There are interdependencies among channels and different channels exhibit distinct characteristics.
    \item \textit{Global Dependencies}: There exist long-term interdependencies and co-evolution among the channels.
    \item \textit{Local Dependencies}: Certain groups of channels exhibit stronger internal dependencies compared to channels outside of these groups.
    \item \textit{Dynamic Dependencies}: The dependencies among channels are dynamic. Even for the same time series, the degree of interaction between channels varies on specific downstream tasks.
\end{itemize}
After identifying the four inductive biases mentioned above, we choose the state space model as the framework for modeling inter-channel dependencies for the following reasons:
\begin{itemize}
    \item As shown in Eq.\ref{eq:ssm1}, the state space model updates through the iterative evolution of the hidden state, representing an adaptive updating mechanism. In contrast to the explicit modeling used in the attention structure, it is more accurate when handling complex, long-term inter-channel interactions. Furthermore, while attention relies on the static attention matrix, its ability to describe dynamic dependencies is inferior to that of the state space models.
    \item Convolution channel modeling suffers from the following limitations: On one hand, the description of local dependencies is influenced by the convolution kernel, which assumes that all dependencies are fixed in space (or time) and cannot capture varying local dependencies. On the other hand, the range of dependencies is constrained by the receptive field, making it incapable of directly and effectively modeling global dependencies.
\end{itemize}
The CMB module takes a latent representation $x\in\mathbb{R}^{B\times C\times L}$ as input. First, the input representation $x$ is transposed to $x^T\in\mathbb{R}^{B\times L\times C}$ for processing along the channel dimension, and the transposed representation is passed through a convolution module and a PNM module. In parallel, a combination of a convolutional layer followed by a sigmoid activation is utilized to dynamically adjust the relative relationships between the channels. After a residual connection with the transposed input $x^T$, the CMB module transposes the output back to the original shape of $x$ to produce the final output. The details of CMB block are shown in Fig.\ref{fig:modules}c. 

\subsection{Diffusion Models for Time Series Imputation}
\label{sec:diff}
When dealing with time series imputation using diffusion models, consider a time series $\tilde{X}$, our goal is to model the posterior $P(\bar{X}\vert X, M, T)$. To make the modeled posterior more precisely, it is natural to introduce conditions to introduce the diffusion process. Considering the short range and long range inter-dependencies within time series, maximizing the observed values utilized in the diffusion process can effectively improve the performance of the imputation results. On the other hand, due to the fact that all the observed values are utilized as condition inputs in the diffusion process, we do not apply any extra process to the observed values to avoid the error accumulation caused by information propagation, the observed values $X_o^c$ are condition inputs for the diffusion process. Thus, the reverse process in Eq.\ref{eq:rprocess:con} is modified to a conditional form with time-series inputs:
\begin{equation}
    p_{\theta}(X^{m}_{0:T}\vert X_0,X_o^c) = p(X_{T}^{m})\prod_{t=1}^{T}p_{\theta}(X_{t-1}^m\vert X_t^{m},X_o^c),
\end{equation}
where $X_T^m\sim \mathcal{N}(0,I)$, $X_t^m$ denotes the sequence of latent variables in the diffusion process and $t\in\{1,2,\cdots,T\}$ is the diffusion time steps. Eq.\ref{eq:rp:con} is reformulated as:
\begin{equation}
    p_{\theta}(X_{t-1}^m\vert X_{t}^m,X_o^c) = \mathcal{N}(X_{t-1}^m;\mu_{\theta}(X_t^m,t\vert X_o^c),\sigma_{\theta}(X_t^m,t\vert X_o^c)\bm{I}),
    \label{eq:condmean}
\end{equation}

Then, the parameterized mean turns to:
\begin{equation}
  \mu_{\theta}(X_t,t) = \frac{1}{\alpha_t}\left(X_t-\frac{\beta_t}{\sqrt{1-\alpha_t}}\epsilon_{\theta}(X_t,t\vert X_o^c)\right),
\end{equation}

\noindent
where 
\begin{equation}
    X_t = \sqrt{\alpha_t}X_0 + (1-\alpha_t)\epsilon,
    \label{eq:xt}
\end{equation}
$\{\beta_t\in(0,1)\}_{t=1}^T$ is a predefined variance scheduler, and $\alpha_t = \prod_{i=1}^t(1-\beta_t)$. 

Finally, we get the conditional diffusion loss for the time series imputation task:
\begin{equation}
  \mathcal{L} = \mathbb{E}_{X_0,\epsilon}\Vert \epsilon-\epsilon_{\theta}(X_t,t\vert X_o^c)\Vert=\mathbb{E}_{X_0,\epsilon}\Vert \epsilon-\epsilon_{\theta}(\sqrt{\alpha_t}X_0 + (1-\alpha_t)\epsilon,t\vert X_o^c)\Vert,
  \label{eq:closs}
\end{equation}
where $\epsilon\sim \mathcal{N}(0,I)$.

The network is trained based on Eq.\ref{eq:closs} to obtain the parameterized noise estimation function $\epsilon_\theta$. And the missing values are estimated by iterative sampling using $\epsilon_\theta$. The training and sampling algorithm is detailed in Alg.\ref{alg:training} and Alg.\ref{alg:inference}.

Compared with the attention-based backbones, the advantage of our proposed backbone lies in the following:
\begin{enumerate}
    \item Our proposed is more capable of distinguishing noise and signal. The data is not always made up of signals over the diffusion process. Self-attention mechanism generates attention weights according to input data, i.e.,$y=\text{softmax}\frac{QK^T}{\sqrt{d}}V$., which means in the early stage of the diffusion process (data is made up of noise totally) and the intermediate stage of diffusion process (data is made up of noise and signal), the attention weights may be misleading. To avoid this issue, we proposed to use linear state space models (which updates $h_t = A_th_{t-1} + B_tx_t$ and outputs $y_t = C_t^Th_t$ or $y=\sum_{k=0}^tK_{t-k}x_k$ equivalently) as the backbone diffusion modules, which update parameters independent of content similarity and the gating mechanism of linear SSMs can also help filter noise.
    \item  Our proposed backbone has a controllable frequency response. From a frequency-domain perspective, SSMs offer controllable frequency responses $\hat{K}(\omega)=C(i\omega I-A)^{-1}B$, allowing them to suppress broadband noise while preserving signal components concentrated in specific frequency bands. This frequency selectivity enables more robust and adaptive representation learning, whereas attention mechanisms, lacking frequency bias, may fail under similar conditions.

\end{enumerate}

\begin{algorithm}[t]
  \caption{Training Procedure of SSD-TS}
  \label{alg:training}
  \begin{algorithmic}[1]
    \STATE \textbf{Input:} Observed sequence $x_0$, number of iterations $N$, variance scheduler $\beta_t$
    \STATE \textbf{Output:} Denoising function $\epsilon_{\theta}$
    \STATE \textbf{For} i = 1 \textbf{to} $N$ \textbf{do}:
    \STATE\quad $t\sim$ \text{Uniform}$(\{1,2,\cdots,T\})$
    \STATE\quad $\epsilon\sim\mathcal{N}(0,I)$
    \STATE\quad Calculate diffusion targets $x_t$ according to Eq.\ref{eq:xt}
    \STATE\quad Take gradient step on
    \begin{equation*}
    \nabla_{\theta}(\Vert\epsilon-\epsilon_{\theta}(x_t,t\vert X_0)\Vert)   
    \end{equation*}
    according to Eq.\ref{eq:closs}
    \STATE \textbf{End For}
  \end{algorithmic}
\end{algorithm}
\begin{algorithm}[t]
  \caption{Sampling Procedure of SSD-TS}
  \label{alg:inference} 
  \begin{algorithmic}[1]
    \STATE\textbf{Input:} Trained denoising function $\epsilon_{\theta}$, sampling step $T$
    \STATE\textbf{Output:} Mean prediction $x_0$
    \STATE\textbf{For} $t = T, T-1, \cdots, 1$ \textbf{do}:
    \STATE\quad $z\sim \mathcal{N}(0,I)$ if $t>1$ else $z=0$
    \STATE\quad $x_{t-1} = \frac{1}{\sqrt{\alpha_t}} \left(x_t-\frac{1-\alpha_t}{\sqrt{1-\bar{\alpha}_t}}\right)\epsilon_{\theta}(x_t,t)+\sigma_tz$
    \STATE\textbf{End For}
  \end{algorithmic}
\end{algorithm}


\subsection{Complexity Analysis}
In this part, we will give a brief analysis about the time and space complexity in the SSM module and self-attention module\footnote{We do not take the time and space complexity of MLPs before the self-attention module or SSM module into consideration.}. While dealing with the input sequences, the core component of our module is the PNM module in Fig.\ref{fig:modules} and the self-attention module in the Transformer architecture, respectively. The time complexity of self-attention module is $O(CL^2)$ and the space complexity is $O(L^2+CL)$, where $L$ is the length of the input sequence and $C$ is the channel of the input sequence. 

In our method, the forward process described in Eq.\ref{eq:disssm} is implemented by converting the process to multiplications of structured matrices, which is of time complexity $O(NCL)$ and of space complexity $O(CL+N(C+L))$ ($N$ is a constant number and set as $16$ by default). This indicates that our model is of linear time and space complexity with respect to the sequence length $L$, which ensures scalability and reduces memory cost for longer sequences.
\section{Experiments}
\label{sec:exp}
\subsection{Experimental Settings and Evaluation Metrics}
All experiments are conducted using PyTorch~\cite{paszke2019pytorch} in Python 3.9 and executed on an NVIDIA RTX3090 GPU. The training process is guided by Eq.\ref{eq:closs}, employing the ADAM optimizer~\cite{DBLP:journals/corr/KingmaB14}  with a learning rate of $2\times 10^{-4}$. The hyperparameters in our experiments can be found in Appendix.\ref{app:hpara}.

For time series modeling tasks, we explicitly categorize the problem into two paradigms based on the model's capacity to quantify uncertainty, \textit{i.e.}, probabilistic time series modeling and deterministic time series modeling. To evaluate the performance of deterministic modeling methods, we report the Mean Absolute Error~(MAE), 
the Mean Squared Error~(MSE), Mean Relative Error~(MRE) and Root Mean Squared Error~(RMSE) between outputs of the model and ground-truth time series data.
As defined in Definition.\ref{def:incts}, the original time series is denoted as $y\in\mathbb{R}^{K\times L}$, the output time series is denoted as $\hat{y}\in\mathbb{R}^{K\times L}$, $M$ is the indicator matrix, the metrics are formulated as follows:

\noindent \textbf{Mean Absolute Error~(MAE)}: MAE calculates the average $L_1$ distance between ground truth and the output values alongside the channel dimension, which is formulated as:
\begin{equation}
    \mathbf{MAE}(y,\hat{y}) = \frac{1}{k}\sum_{i=1}^{K}\sum_{j=1}^{L}\vert(y-\hat{y})\odot(1-M)\vert_{i,j}
\end{equation}

\noindent\textbf{Mean Square Error~(MSE)}: MSE calculates the average $L_2$ between ground truth and the output values alongside the channel dimension, which is formulated as:
\begin{equation}
    \mathbf{MSE}(y,\hat{y}) = \frac{1}{k}\sum_{i=1}^{K}\sum_{j=1}^{L}((y-\hat{y})\odot(1-M))^2_{i,j}
\end{equation}

\noindent\textbf{Root Mean Square Error~(RMSE)}: RMSE is the square root of the average $L_2$ between ground truth and the output values alongside the channel dimension:
\begin{equation}
    \begin{aligned}
        \mathbf{RMSE}(y,\hat{y}) &= \sqrt{\frac{1}{k}\sum_{i=1}^{K}\sum_{j=1}^{L}((y-\hat{y})\odot(1-M))^2_{i,j}}
    \end{aligned} 
\end{equation}

\noindent\textbf{Mean Relative Error~(MRE)}: MRE estimates the relative difference between $y$ and $\hat{y}$:
\begin{equation}
    \mathbf{MRE}(y,\hat{y}) = \frac{1}{k}\sum_{i=1}^{K}\sum_{j=1}^{L}(1-M)_{i,j}\odot\frac{\vert(y-\hat{y})\vert_{i,j}}{y_{i,j}}
\end{equation}

For probabilistic time series modeling evaluation, we additionally evaluate the Continuous Ranked Probability Score (CRPS). 
CRPS measures the integral squared difference between the predicted and empirical cumulative distribution functions (CDFs), providing a pointwise evaluation of probabilistic calibration. 
By applying CRPS, we can effectively evaluate the consistency between the estimated distribution and the real distribution. The CRPS metric is defined as follows:

\noindent\textbf{Continuous Ranked Probabilistic Score~(CRPS)}: Given an estimated probability distribution function $F$ modeled with an observation $x$, CRPS evaluates the compatibility and is defined as the integral of the quantile loss for all quantile levels:
\begin{equation}
    \mathbf{CRPS}(F^{-1},x) = \int_{0}^1\Lambda_{\alpha}(F^{-1}(\alpha,x))\ \text{d}\alpha,
\end{equation}
where $\Lambda_{\alpha}(q,y) = (\alpha-\bm{1}_{y<q})(y-q),\alpha\in[0,1]$ and $\bm{1}_{y<q}$ the indicator function, \textit{i.e.}, if $y<q$, the value of the indicator function is $1$, else $0$.

Following \cite{DBLP:conf/nips/TashiroSSE21,DBLP:conf/icml/YanGHZX24}, we separate the interval $[0,1]$ to $20$ quantile levels with a stepsize of $s=0.05$, and the estimated value of CRPS is:
\begin{equation}
    \mathbf{CRPS}(F^{-1},x)\approx \sum_{i=1}^{19}\frac{2\Lambda_{i\cdot s}(F^{-1}(i\cdot s,x))}{19}
\end{equation}
For the whole time series $X\in\mathbb{R}^{K\times L}$, the CRPS value is normalized for all time steps and channels:
\begin{equation}
    \mathbf{CRPS}(F^{-1},X) = \frac{\sum_{i=1}^K\sum_{j=1}^{L}\mathbf{CRPS}(F^{-1}_{i,j},X_{i,j})}{\sum_{i=1}^K\sum_{j=1}^{L}\vert X_{i,j}\vert}
\end{equation}

\subsection{Time Series Imputation}
\subsubsection{Datasets}
To evaluate the performance of our model on time series imputation tasks, we employ three benchmark datasets, including the MuJoCo dataset~\cite{DBLP:conf/nips/RubanovaCD19}, the Air Quality dataset~(AQI)~\cite{DBLP:conf/ijcai/YiZZL16} and Physionet Challenge 2012 dataset~\cite{silva2012predicting}. Please refer to Appendix.\ref{app:datadetail} for more details about the datasets.

It is worth noting that we select these three datasets to include as many potential missing data scenarios as possible. For the Physionet dataset, consistent with previous studies~\cite{DBLP:conf/nips/CaoWLZLL18,suo2020glima,qiu2024tfb,qiu2025tab}, we process the dataset into hourly time series, each containing 48 time steps, with approximately 80\% of the data missing. Since the dataset does not include ground truth, we randomly select 10\%, 50\%, and 90\% of the data as ground truth in the test set. For the Air Quality dataset, around 13\% of the data is missing in a non-random manner. As this dataset contains structured artificial ground truth, we do not test different missing ratios on the AQI dataset.
The MuJoCo dataset, on the other hand, is a complete dataset and we apply hand-crafted masks to simulate masking patterns that are not present in the Physionet and Air Quality datasets.



\subsubsection{Baselines}
Correspondingly, in our experiments, the baselines include both deterministic and probabilistic imputation models. The deterministic models involve: BRITS~\cite{DBLP:conf/nips/CaoWLZLL18}, RDIS~\cite{DBLP:journals/access/ChoiKK23a}, SSGAN~\cite{DBLP:conf/aaai/MiaoWWGMY21}, TIDER~\cite{DBLP:conf/iclr/LiuLCCJ23}, and SAITS~\cite{DBLP:journals/eswa/DuCL23}. The probabilistic imputation models encompass various distribution estimation methods, such as: methods based on variational autoencoders~(V-RIN~\cite{DBLP:journals/tcyb/MulyadiJS22}), methods based on diffusion models~(CSDI~\cite{DBLP:conf/nips/TashiroSSE21}, SSSD~\cite{DBLP:journals/tmlr/AlcarazS23}, D$^3$M\cite{DBLP:conf/icml/YanGHZX24}, TS-Diff~\cite{DBLP:conf/iclr/LiuLCCJ23}), methods based on Schr\"{o}dinger Bridge~(CSBI~\cite{DBLP:conf/icml/ChenDFLYZRZSN23}), and methods based on Gaussian process~(GP-VAE~\cite{DBLP:conf/aistats/FortuinBRM20}). And on the MuJoCo dataset, we compare our models against RNN GRU-D~\cite{che2018recurrent}, ODE-RNN~\cite{DBLP:conf/nips/RubanovaCD19}, NeuralCDE~\cite{DBLP:conf/icml/MorrillSKF21}, Latent-ODE~\cite{DBLP:conf/nips/RubanovaCD19}, NAOMI~\cite{DBLP:conf/nips/MariscaCA22}, CSDI~\cite{DBLP:conf/nips/TashiroSSE21} and SSSD~\cite{DBLP:journals/tmlr/AlcarazS23} to be consistent with SSSD~\cite{DBLP:journals/tmlr/AlcarazS23}.
\subsubsection{Experimental Results and Analysis} 
Table \ref{tab:mujoco} presents the experimental results on the MuJoCo dataset for the random missing~(RM) case, with different missing rate scenarios. Under the two higher missing rates (80\% and 90\%), our method achieves the best performance, with improvements of 48.2\% (at 80\% missing rate) and 65.8\% (at 90\% missing rate) compared to the second-best results. At 70\% missing rate, our method ranks second, with only an 11.1\% difference from the best-performing CSDI. This demonstrates the strong modeling capability of our method on datasets with high missing rates under random missing scenario.
\begin{table}[tbp]
\centering
\caption{MSE Results on MuJoCo Dataset with missing ratio 70$\%$, 80$\%$ and 90$\%$ for the missing scenario \textit{RM}.}
\label{tab:mujoco}
\resizebox{\linewidth}{!}{
\begin{tabular}{l|ccc}
\hline
\multicolumn{1}{c|}{\textbf{Model}} & \textbf{70$\%$ RM}   & \textbf{80$\%$ RM}   & \textbf{90$\%$ RM}   \\ \hline
RNN GRU-D                           & 1.134e-2              & 1.421e-2              & 1.968e-2              \\
ODE-RNN                             & 9.86e-3              & 1.209e-2              & 1.647e-2              \\
NeuralCDE                           & 8.35e-3              & 1.071e-2             & 1.352e-2              \\
Latent-ODE                          & 3.00e-3              & 2.95e-3              & 3.60e-3              \\
NAOMI                               & 1.46e-3              & 2.32e-3              & 4.42e-3              \\
NRTSI                               & 6.3e-4              & 1.22e-3              & 4.06e-3              \\
CSDI                                & \textbf{2.4e-4$\pm$3e-5}  & \underline{6.1e-4$\pm$1.0e-4}  & 4.84e-3$\pm$2e-5  \\
SSSD                                & 5.9e-4$\pm$8e-5  & 1e-3$\pm$5e-5  & \underline{1.90e-3$\pm$3e-5}  \\
SSD-TS(Ours)                      & \underline{2.7e-4$\pm$1e-5} & \textbf{3.16e-4$\pm$9.77e-6} &  \textbf{6.5e-4$\pm$1e-4} \\ \hline
\end{tabular}}
\end{table}

Table.\ref{tab:rmse_imp} presents the RMSE performance of our method on two real-world datasets with missing data. We observe that on the Physionet dataset, our method achieves the best results across three different missing rates (10\%, 50\%, and 90\%), with improvements of 22.6\% (at 10\% missing rate), 17.2\% (at 50\% missing rate), and 23.5\% (at 90\% missing rate). Additionally, on the AQI dataset, our method ranks second. These results demonstrate that our method outperforms other approaches in modeling time series data on real-world datasets with missing values.
\begin{table}[t]
\caption{RMSE results on Physionet and Air quality dataset. The mean and standard error are obtained by 3 runs and the best result are in \textbf{bold} while the second best results are \underline{underlined}.}
\label{tab:rmse_imp}
\resizebox{\linewidth}{!}{
\begin{tabular}{c|ccc|c}
\hline
               & \multicolumn{3}{c|}{Physionet}                                                                                                    & AQI                        \\ \cline{2-4} 
               & \multicolumn{1}{c|}{10\% missing}                & \multicolumn{1}{c|}{50\% missing}                & 90\% missing                &                            \\ \hline
V-RIN          & \multicolumn{1}{c|}{0.628$\pm$0.025}             & \multicolumn{1}{c|}{0.693$\pm$0.022}             & 0.928$\pm$0.013             & 40.11$\pm$1.14             \\
BRITS          & \multicolumn{1}{c|}{0.619$\pm$0.018}             & \multicolumn{1}{c|}{0.701$\pm$0.021}             & 0.847$\pm$0.021             & 24.28$\pm$0.65             \\
SSGAN          & \multicolumn{1}{c|}{0.607$\pm$0.034}             & \multicolumn{1}{c|}{0.758 $\pm$0.025}            & 0.830$\pm$0.009             & -                          \\
RDIS           & \multicolumn{1}{c|}{0.635$\pm$0.018}             & \multicolumn{1}{c|}{0.747 $\pm$0.013}            & 0.922$\pm$0.018             & 37.25$\pm$0.31             \\
CSDI           & \multicolumn{1}{c|}{0.531$\pm$0.009}             & \multicolumn{1}{c|}{0.668$\pm$0.007}             & 0.834$\pm$0.006             & 19.21$\pm$0.13             \\
CSBI           & \multicolumn{1}{c|}{0.547$\pm$0.019}             & \multicolumn{1}{c|}{0.649 $\pm$0.009}            & 0.837$\pm$0.012             & 19.07$\pm$0.18             \\
SSSD           & \multicolumn{1}{c|}{0.459$\pm$0.001}             & \multicolumn{1}{c|}{0.632$\pm$0.004}             & 0.824$\pm$0.003             & 18.77$\pm$0.08             \\
TS-Diff        & \multicolumn{1}{c|}{0.523$\pm$0.015}             & \multicolumn{1}{c|}{0.679$\pm$0.009}             & 0.845$\pm$0.007             & 19.06$\pm$0.14             \\
SAITS          & \multicolumn{1}{c|}{0.461$\pm$0.009}             & \multicolumn{1}{c|}{0.636$\pm$0.005}             & 0.819$\pm$0.002             & 18.68$\pm$0.13             \\
D$^3$M(Constant-Sqrt)         & \multicolumn{1}{c|}{\underline{0.438$\pm$0.003}} & \multicolumn{1}{c|}{\underline{0.615$\pm$0.012}} & \underline{0.814$\pm$0.002} & \textbf{18.19$\pm$0.18}    \\
TIDER          & \multicolumn{1}{c|}{0.486$\pm$0.006}             & \multicolumn{1}{c|}{0.659$\pm$0.009}             & 0.833$\pm$0.005             & 18.94$\pm$0.21             \\
SSD-TS(Ours) & \multicolumn{1}{c|}{\textbf{0.339$\pm$0.0002}}   & \multicolumn{1}{c|}{\textbf{0.509$\pm$0.007}}    & \textbf{0.623$\pm$0.0001}   & \underline{18.66$\pm$0.26} \\ \hline
\end{tabular}}
\end{table}

Table.\ref{tab:crps} presents the CRPS performance of our method on two real-world datasets with missing data. Compared to other probabilistic time series methods, our approach achieves the lowest CRPS scores in three out of the four tasks (10\% and 50\% on PhysioNet, and AQI). This indicates that, relative to other probabilistic time series modeling methods, our approach demonstrates the most accurate modeling capability of the distribution of unknown points within the sequence.
\begin{table}[t]
\caption{CRPS results on Physionet and Air quality dataset. The mean and standard error are obtained by 3 runs and the best result are in \textbf{bold} while the second best results are \underline{underlined}.}
\label{tab:crps}
\resizebox{\linewidth}{!}{
\begin{tabular}{c|ccc|c}
\hline
                      & \multicolumn{3}{c|}{Physionet}                                                                                              & AQI                       \\ \cline{2-4}
                      & \multicolumn{1}{c|}{10\% missing}              & \multicolumn{1}{c|}{50\% missing}              & 90\% missing              &                           \\ \hline
GP-VAE                & \multicolumn{1}{c|}{0.582$\pm$0.003}           & \multicolumn{1}{c|}{0.796$\pm$0.004}           & 0.998$\pm$0.001           & 0.402$\pm$0.009           \\
V-RIN                 & \multicolumn{1}{c|}{0.814$\pm$0.004}           & \multicolumn{1}{c|}{0.845$\pm$0.002}           & 0.932$\pm$0.001           & 0.534$\pm$0.013           \\
CSDI                  & \multicolumn{1}{c|}{0.242$\pm$0.001}           & \multicolumn{1}{c|}{0.336$\pm$0.002}           & 0.528$\pm$0.003           & 0.108$\pm$0.001           \\
CSBI                  & \multicolumn{1}{c|}{0.247$\pm$0.003}           & \multicolumn{1}{c|}{0.332 $\pm$0.003}          & 0.527$\pm$0.006           & 0.110$\pm$0.002           \\
SSSD                  & \multicolumn{1}{c|}{0.233$\pm$0.001}           & \multicolumn{1}{c|}{0.331$\pm$0.002}           & \underline{0.522$\pm$0.002} & 0.107$\pm$0.001           \\
TS-Diff               & \multicolumn{1}{c|}{0.249$\pm$0.002}           & \multicolumn{1}{c|}{0.348$\pm$0.004}           & 0.541$\pm$0.006           & 0.118$\pm$0.003           \\
D$^3$M(Constant-Sqrt) & \multicolumn{1}{c|}{\underline{0.223$\pm$0.001}} & \multicolumn{1}{c|}{\underline{0.327$\pm$0.003}} & \textbf{0.520$\pm$0.001}  & \underline{0.106$\pm$0.002} \\
SSD-TS(Ours)          & \multicolumn{1}{c|}{\textbf{0.164$\pm$0.0004}} & \multicolumn{1}{c|}{\textbf{0.244$\pm$0.0001}} & 0.533$\pm$0.0004          & \textbf{0.096$\pm$0.0002} \\ \hline
\end{tabular}}
\end{table}

In addition, we also compare the performance of diffusion models with different backbones. Existing imputation methods mainly cover two-kinds of backbones: transformer-based~\cite{DBLP:conf/nips/VaswaniSPUJGKP17} backbones and state-space model-based~\cite{DBLP:conf/iclr/GuGR22} backbones.
The core architecture of CSDI is a spatial-temporal transformer, which first embeds the input tokens and captures the inter- and intra- channel dependency with two transformer encoders. D$^3$M adopts a similar attention mechanism as the backbone, with the addition of gating mechanisms and an exponential moving average~(EMA) module.
SSSD modifies the backbone of CSDI by replacing the spatial-temporal transformer with the S4 structure proposed in \cite{DBLP:conf/iclr/GuGR22}, which is also utilized in TS-Diff.

\begin{table*}[tbp]
\centering
\caption{Experimental Results of Ablation Study}
\label{tab:abl}
\resizebox{\linewidth}{!}{
\begin{tabular}{c|c|c|l|l|l|l}
\hline
Time Modeling                      & Temporal Attention & Inter-Channel Dependency & \multicolumn{1}{c|}{MSE}     & \multicolumn{1}{c|}{MAE}     & \multicolumn{1}{c|}{MRE}    & \multicolumn{1}{c}{RMSE}    \\ \hline
Bidirectional                      & Yes                & Yes                      & \textbf{5.46e-4$\pm$1.6e-5}           & \textbf{1.17e-2$\pm$7.4e-5}           & \textbf{1.21$\pm$7.5e-3\%}           & \textbf{2.33e-2$\pm$3.1e-4}           \\
Forward                            & Yes                & Yes                      & \underline{7.19e-4$\pm$2.0e-5} & 1.26e-2$\pm$2.1e-4           & 1.29$\pm$2.2e-2\%           & \underline{2.67e-2$\pm$2.9e-4} \\
Forward                            & Yes                & No                       & 7.48e-4$\pm$9.5e-5           & \underline{1.23e-2$\pm$3.5e-4} & \underline{1.23$\pm$3.5e-2\%} & 2.71e-2$\pm$1.5e-3           \\
Backward                           & Yes                & Yes                      & 7.24e-4$\pm$7.3e-5           & 1.30e-2$\pm$4.2e-4           & 1.30$\pm$4.2e-2\%           & 2.69e-2$\pm$1.3e-3           \\
Backward                           & Yes                & No                       & 8.39e-4$\pm$6.1e-5           & 1.46e-2$\pm$3.8e-4           & 1.46$\pm$3.8e-2\%           & 2.89e-2$\pm$1.0e-3           \\
Bidirectional                      & Yes                & No                       & 8.85e-4$\pm$2.8e-5           & 1.40e-2$\pm$3.1e-4           & 1.44$\pm$3.4e-2\%           & 2.97e-2$\pm$4.8e-4           \\
Bidirectional                      & No                 & Yes                      & 9.66e-4$\pm$9.5e-5           & 1.53e-2$\pm$3.3e-4           & 1.57$\pm$3.5e-2\%           & 3.09e-2$\pm$1.3e-3           \\
Bidirectional & Yes & Channel Attention     & 7.43e-4$\pm$4.0e-5           & 1.31e-2$\pm$5.5e-5           & 1.35$\pm$5.6e-3\%           & 2.71e-2$\pm$5.6e-5           \\ \hline
\end{tabular}}
\end{table*}

As stated in Sec.\ref{sec:method}, our method employs a modern state space model, Mamba~\cite{DBLP:conf/icml/DaoG24} as the backbone. Similar to other methods, we also adopt special designs to capture inter- and intra- channel dependencies. The results in Table.\ref{tab:rmse_imp} and \ref{tab:crps} show that our method outperforms those with transformer and S4-based core architectures. This demonstrates: 1) the effectiveness of Mamba structure as the backbone for diffusion models and 2) the validity of the designs we adopted in addressing the problem of time series imputation.

\subsection{Visualization Results}
\begin{figure}
    \includegraphics[width=\linewidth]{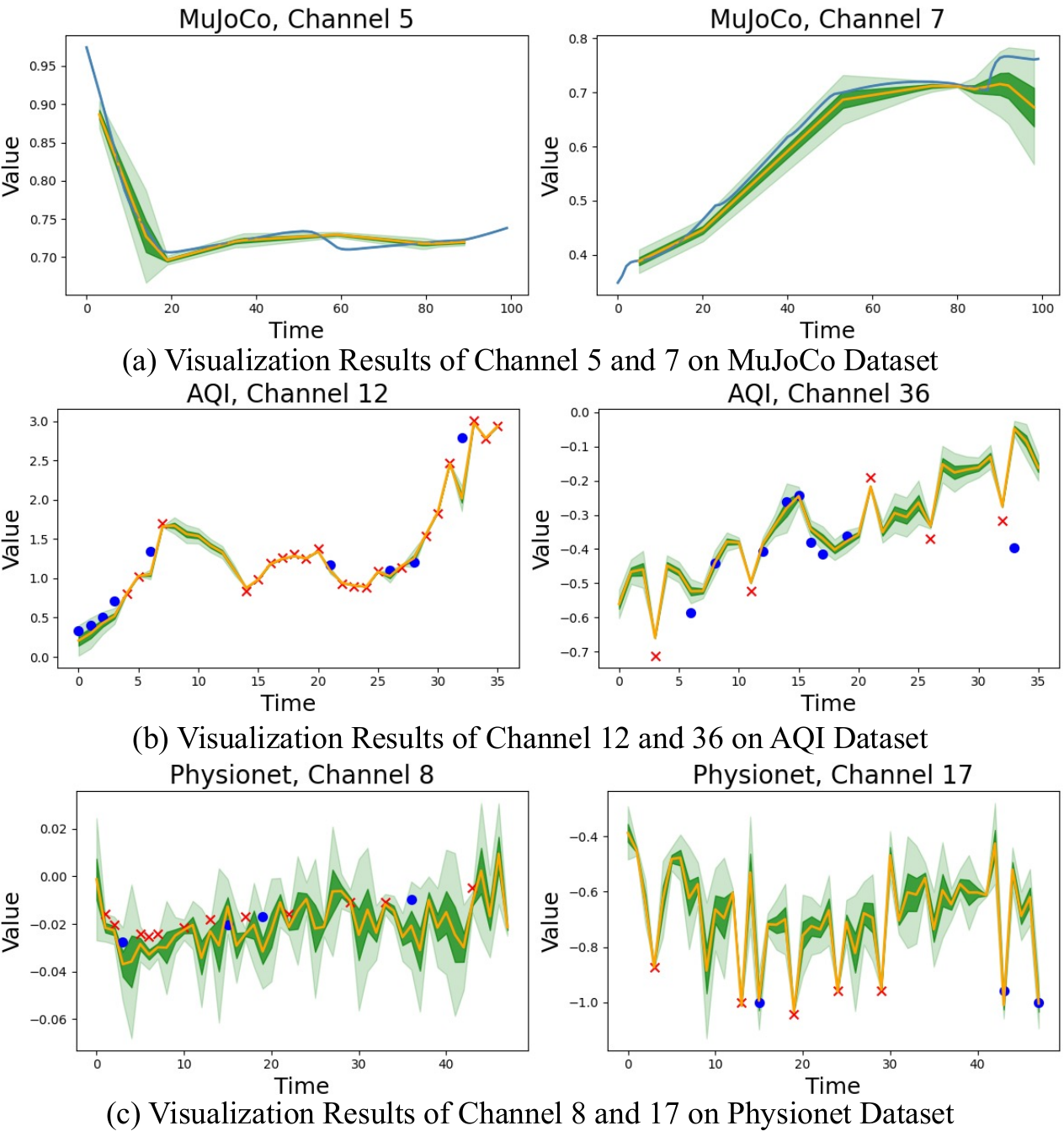}
    \caption{Visualization results of probabilistic time series imputation results on MuJoCo, AQI and Physionet dataset. The orange solid line represents the mean of the imputation results, while the dark green represents the 50\% confidence interval, and the light green solid line represents the 95\% confidence interval. In Figure a, the blue solid line represents the target of imputation (i.e., the ground truth). In Figures b and c, the blue dots represent the ground truth of the missing points, and the red crosses represent the observed values.}
    \label{fig:visual}
\end{figure}
Fig.\ref{fig:visual} shows the visualization results on the MuJoCo dataset, AQI dataset and Physionet dataset. We can see that almost all ground truth values for the points to be imputed fall within the 95\% confidence interval, and most of the ground truth values are within the 50\% confidence interval, which demonstrates the effectiveness of our method. 

\subsection{Ablation Studies}

\textbf{Effectiveness of Proposed Modules.} To validate the effectiveness of the proposed module, we conduct ablation experiments on the following aspects: 1)~the bidirectional modeling, 2)~the temporal attention mechanism, 3)~the inter-channel multivariate dependencies. We also replace the CMB block with channel attention module implemented using \cite{DBLP:conf/cvpr/HuSS18} to validate the effectiveness of CMB block. All experiments are conducted on the MuJoCo dataset with the missing ratio 90\%. 

The results are shown in Table.\ref{tab:abl}. It can be observed that the module equipped with BAM and CMB block performs the best, significantly outperforming the results of removing any one of these components across all four metrics. The temporal attention module has the largest impact on the model, and its removal leads to a significant performance drop. Similarly, removing the CMB module also results in a notable degradation in performance, meanwhile, modifying the CMB block to Channel Attention in \cite{DBLP:conf/cvpr/HuSS18} also leads to a performance drop, which proves that state space models are also effective for modeling inter-channel dependencies. On the other hand, adjusting the BAM module to its unidirectional form also causes some degree of performance decrease, which means the bidirectional modeling manner is capable of modeling more intra-channel dependency. This fully demonstrates the effectiveness of all our proposed blocks and the capability of linear state space models as backbones for time series diffusion models.

\textbf{Time Efficiency of Mamba Backbones.} We compare the inference time of our model with a variant using the transformer backbone. The results is shown in Table.\ref{tab:timecmp}. We can see that our model achieves faster sampling, less memory cost comparing with the model with transformer backbones, which demonstrates the efficiency of our model and our proposed backbone.

\begin{table}[htbp]
\caption{Inference time comparison with transformer backbones}
\label{tab:timecmp}
\begin{tabular}{c|c|c}
\hline
Model                    & Transformer-backbone & SSD-TS(128) \\ \hline
\#Parameters(M)           & 9.208                 & 87.57       \\
Inference time/sample(s) & 1.5126                & 0.93        \\
GPU Memory(MB)           & 11250                 & 4536        \\ \hline
\end{tabular}
\end{table}

\subsection{Parameter Sensitivity Analysis}
We conduct parameter sensitivity experiment and the results are shown in Table.\ref{tab:params}.
There are three hyperparameters with different dimensions: sequence, residual connection, and input projection dimension, which are set to be equal in our experiments. We test different results for $C = 32,64,128$. The experimental results show that as $C$ increases, all metrics significantly decrease. Additionally, since $C = 256$ exceeds one single GPU memory capacity, we choose $C = 128$ in our experiments to balance between metrics and computational cost.
\begin{table}[htbp]
\caption{Parameter Sensitivity Results}
\label{tab:params}
\begin{tabular}{c|c|c|c}
\hline
\#Channel & MAE    & MSE     & RMSE   \\ \hline
32       & 0.0482 & 0.0066  & 0.0809 \\
64       & 0.0147 & 0.00075 & 0.0273 \\
128      & 0.0135 & 0.00065 & 0.0254 \\ \hline
\end{tabular}
\end{table}
\section{Related Work}
\label{sec:rel}
\subsection{Time Series Imputation}
The objective of time series imputation is to recover the missing data points leveraging the observed values and intrinsic dependencies within the series. Statistical time series methods, such as autoregressive~(AR) models~\cite{meek2002autoregressive,fuller1981properties} and autoregressive integrated moving average~(ARIMA) models~\cite{newbold1983arima,nelson1998time}, are capable of imputing missing values; however, they often lack the flexibility required to accurately model the complex characteristics of time series data, such as trends and seasonality. Deep learning methods have also found extensive application in imputation tasks. From the perspective of network architectures, various structures including TCN~\cite{DBLP:conf/cvpr/LeaFVRH17}, LSTM~\cite{DBLP:journals/neco/HochreiterS97}, Multi-Layer Perceptron~(MLP), and Transformer~\cite{DBLP:conf/nips/VaswaniSPUJGKP17} have been applied to time series imputation, such as \cite{DBLP:journals/eswa/DuCL23,DBLP:conf/cikm/ZhangLY23, 9964035, DBLP:conf/huc/YuanXYJZ18, DBLP:journals/nca/ParkMAFPVSA23,chen2024pathformer}. State space models have also been applied to time series imputation tasks, such as \cite{DBLP:journals/corr/abs-2405-16312,DBLP:conf/iclr/DinhSB17}. Recently, methods based on generative methods~\cite{DBLP:conf/nips/TashiroSSE21,DBLP:journals/tmlr/AlcarazS23,DBLP:conf/icml/YanGHZX24} have also achieved great success in probabilistic time series imputation. 
\subsection{Generative Models}
Generative models are probabilistic models that learn the original data distribution and generate new samples by sampling from the learned distribution. The goal is to learn the joint probability distribution $p(x)$ of the training data or the posterior distribution $p(x\vert y)$ and generate new samples that resemble the original data. Traditional generative models include Hidden Markov Models~\cite{DBLP:journals/corr/abs-2310-14341,DBLP:conf/cvpr/LiW0021}, Bayesian models~\cite{DBLP:conf/icml/FangWLZ024,DBLP:journals/eswa/KocadagliA14} and Gaussian process~\cite{DBLP:conf/pkdd/CoraniBZ21,DBLP:journals/access/ZhangWPQLFW21}. Researchers have also proposed many deep learning-based generative models, such as Variational Autoencoders~\cite{DBLP:journals/corr/KingmaW13}, Generative Adversarial Networks (GANs)~\cite{DBLP:conf/nips/GoodfellowPMXWOCB14}, Normalizing Flow-based Models~\cite{DBLP:journals/pami/KobyzevPB21}, and Denoising Diffusion Probabilistic Models~\cite{DBLP:conf/nips/HoJA20}.

Since the goal of generative models is to model the distribution of unknown points and generate target points by sampling from the learned distribution, they naturally possess the ability to estimate the uncertainty of unknown points. This makes generative models particularly well-suited for probabilistic time series modeling. Therefore, many methods have been proposed which adopts generative models for probabilistic time series modeling, including VAE-based methods~\cite{DBLP:journals/corr/abs-2111-08095,DBLP:journals/tnn/LiYWJ21,wu2025k2vae}, GAN-based methods~\cite{DBLP:conf/nips/YoonJS19,DBLP:conf/nips/LuoCZXY18}, Normalizing-flow based methods~\cite{DBLP:conf/iclr/RasulSSBV21,DBLP:conf/iclr/DaiC22} and diffusion-based methods~\cite{DBLP:conf/icml/RasulSSV21,DBLP:conf/iclr/ShenCK24}. State-space models have proved their effectiveness as diffusion backbones~\cite{DBLP:journals/tmlr/AlcarazS23,DBLP:conf/icml/YanGHZX24}. 
\section{Conclusion}
\label{sec:con}
In this paper, we propose SSD-TS, which explores the potential of linear state-space model (Mamba) as the backbone structure for diffusion models. Based on the requirements of time series modeling task and the characteristics of time series data, we propose a denoising module for diffusion models with Mamba backbone. The proposed module enables bidirectional structural modeling of sequences and effectively leverages both inter-channel and intra-channel dependencies. Experimental results show that, compared to models with structured state-space models, transformers, and convolutional backbones, our method achieves superior performance in time series imputation task. This demonstrates the capability of the linear state-space Mamba model as a backbone for diffusion models. Moreover, our method achieves favorable results in terms of different deterministic and probabilistic metrics, further validating the effectiveness of our design for time series imputation task.
\begin{acks}
This work was partially supported by National Natural Science Foundation of China (62472174). Jilin Hu is the corresponding author of the work. We gratefully acknowledge the support of the ECNU Multifunctional Platform for Innovation (001) for providing computational resources.
\end{acks}

\clearpage
\bibliographystyle{ACM-Reference-Format}
\bibliography{acmart}


\begin{thebibliography}{93}


\ifx \showCODEN    \undefined \def \showCODEN     #1{\unskip}     \fi
\ifx \showISBNx    \undefined \def \showISBNx     #1{\unskip}     \fi
\ifx \showISBNxiii \undefined \def \showISBNxiii  #1{\unskip}     \fi
\ifx \showISSN     \undefined \def \showISSN      #1{\unskip}     \fi
\ifx \showLCCN     \undefined \def \showLCCN      #1{\unskip}     \fi
\ifx \shownote     \undefined \def \shownote      #1{#1}          \fi
\ifx \showarticletitle \undefined \def \showarticletitle #1{#1}   \fi
\ifx \showURL      \undefined \def \showURL       {\relax}        \fi
\providecommand\bibfield[2]{#2}
\providecommand\bibinfo[2]{#2}
\providecommand\natexlab[1]{#1}
\providecommand\showeprint[2][]{arXiv:#2}

\bibitem[Alcaraz and Strodthoff(2023)]%
        {DBLP:journals/tmlr/AlcarazS23}
\bibfield{author}{\bibinfo{person}{Juan Miguel~Lopez Alcaraz} {and} \bibinfo{person}{Nils Strodthoff}.} \bibinfo{year}{2023}\natexlab{}.
\newblock \showarticletitle{Diffusion-based Time Series Imputation and Forecasting with Structured State Space Models}.
\newblock \bibinfo{journal}{\emph{Trans. Mach. Learn. Res.}}  \bibinfo{volume}{2023} (\bibinfo{year}{2023}).
\newblock


\bibitem[Bai et~al\mbox{.}(2020)]%
        {DBLP:conf/ijcai/0001CW0H20}
\bibfield{author}{\bibinfo{person}{Lu Bai}, \bibinfo{person}{Lixin Cui}, \bibinfo{person}{Yue Wang}, \bibinfo{person}{Yuhang Jiao}, {and} \bibinfo{person}{Edwin~R. Hancock}.} \bibinfo{year}{2020}\natexlab{}.
\newblock \bibinfo{title}{A Quantum-inspired Entropic Kernel for Multiple Financial Time Series Analysis}.
\newblock \bibinfo{numpages}{4453--4460}~pages.
\newblock


\bibitem[Cao et~al\mbox{.}(2018)]%
        {DBLP:conf/nips/CaoWLZLL18}
\bibfield{author}{\bibinfo{person}{Wei Cao}, \bibinfo{person}{Dong Wang}, \bibinfo{person}{Jian Li}, \bibinfo{person}{Hao Zhou}, \bibinfo{person}{Lei Li}, {and} \bibinfo{person}{Yitan Li}.} \bibinfo{year}{2018}\natexlab{}.
\newblock \showarticletitle{{BRITS:} Bidirectional Recurrent Imputation for Time Series}. In \bibinfo{booktitle}{\emph{Advances in Neural Information Processing Systems 31: Annual Conference on Neural Information Processing Systems 2018, NeurIPS 2018}}. \bibinfo{pages}{6776--6786}.
\newblock


\bibitem[Che et~al\mbox{.}(2018)]%
        {che2018recurrent}
\bibfield{author}{\bibinfo{person}{Zhengping Che}, \bibinfo{person}{Sanjay Purushotham}, \bibinfo{person}{Kyunghyun Cho}, \bibinfo{person}{David Sontag}, {and} \bibinfo{person}{Yan Liu}.} \bibinfo{year}{2018}\natexlab{}.
\newblock \showarticletitle{Recurrent neural networks for multivariate time series with missing values}.
\newblock \bibinfo{journal}{\emph{Scientific reports}} \bibinfo{volume}{8}, \bibinfo{number}{1} (\bibinfo{year}{2018}), \bibinfo{pages}{6085}.
\newblock


\bibitem[Chen et~al\mbox{.}(2024)]%
        {chen2024pathformer}
\bibfield{author}{\bibinfo{person}{Peng Chen}, \bibinfo{person}{Yingying Zhang}, \bibinfo{person}{Yunyao Cheng}, \bibinfo{person}{Yang Shu}, \bibinfo{person}{Yihang Wang}, \bibinfo{person}{Qingsong Wen}, \bibinfo{person}{Bin Yang}, {and} \bibinfo{person}{Chenjuan Guo}.} \bibinfo{year}{2024}\natexlab{}.
\newblock \showarticletitle{Pathformer: Multi-scale transformers with adaptive pathways for time series forecasting}.
\newblock \bibinfo{journal}{\emph{arXiv preprint arXiv:2402.05956}} (\bibinfo{year}{2024}).
\newblock


\bibitem[Chen et~al\mbox{.}(2023)]%
        {DBLP:conf/icml/ChenDFLYZRZSN23}
\bibfield{author}{\bibinfo{person}{Yu Chen}, \bibinfo{person}{Wei Deng}, \bibinfo{person}{Shikai Fang}, \bibinfo{person}{Fengpei Li}, \bibinfo{person}{Nicole~Tianjiao Yang}, \bibinfo{person}{Yikai Zhang}, \bibinfo{person}{Kashif Rasul}, \bibinfo{person}{Shandian Zhe}, \bibinfo{person}{Anderson Schneider}, {and} \bibinfo{person}{Yuriy Nevmyvaka}.} \bibinfo{year}{2023}\natexlab{}.
\newblock \bibinfo{title}{Provably Convergent Schr{\"{o}}dinger Bridge with Applications to Probabilistic Time Series Imputation}.
\newblock \bibinfo{numpages}{4485--4513}~pages.
\newblock


\bibitem[Choi et~al\mbox{.}(2023)]%
        {DBLP:journals/access/ChoiKK23a}
\bibfield{author}{\bibinfo{person}{Tae{-}Min Choi}, \bibinfo{person}{Ji{-}Su Kang}, {and} \bibinfo{person}{Jong{-}Hwan Kim}.} \bibinfo{year}{2023}\natexlab{}.
\newblock \showarticletitle{{RDIS:} Random Drop Imputation With Self-Training for Incomplete Time Series Data}.
\newblock \bibinfo{journal}{\emph{{IEEE} Access}}  \bibinfo{volume}{11} (\bibinfo{year}{2023}), \bibinfo{pages}{100720--100728}.
\newblock


\bibitem[Cini et~al\mbox{.}(2022)]%
        {DBLP:conf/iclr/CiniMA22}
\bibfield{author}{\bibinfo{person}{Andrea Cini}, \bibinfo{person}{Ivan Marisca}, {and} \bibinfo{person}{Cesare Alippi}.} \bibinfo{year}{2022}\natexlab{}.
\newblock \showarticletitle{Filling the G{\_}ap{\_}s: Multivariate Time Series Imputation by Graph Neural Networks}. In \bibinfo{booktitle}{\emph{The Tenth International Conference on Learning Representations, {ICLR} 2022}}. \bibinfo{publisher}{OpenReview.net}.
\newblock


\bibitem[Cirstea et~al\mbox{.}(2022)]%
        {DBLP:conf/ijcai/CirsteaG0KDP22}
\bibfield{author}{\bibinfo{person}{Razvan{-}Gabriel Cirstea}, \bibinfo{person}{Chenjuan Guo}, \bibinfo{person}{Bin Yang}, \bibinfo{person}{Tung Kieu}, \bibinfo{person}{Xuanyi Dong}, {and} \bibinfo{person}{Shirui Pan}.} \bibinfo{year}{2022}\natexlab{}.
\newblock \showarticletitle{Triformer: Triangular, Variable-Specific Attentions for Long Sequence Multivariate Time Series Forecasting}. In \bibinfo{booktitle}{\emph{Proceedings of the Thirty-First International Joint Conference on Artificial Intelligence, {IJCAI} 2022}}. \bibinfo{pages}{1994--2001}.
\newblock


\bibitem[Corani et~al\mbox{.}(2021)]%
        {DBLP:conf/pkdd/CoraniBZ21}
\bibfield{author}{\bibinfo{person}{Giorgio Corani}, \bibinfo{person}{Alessio Benavoli}, {and} \bibinfo{person}{Marco Zaffalon}.} \bibinfo{year}{2021}\natexlab{}.
\newblock \showarticletitle{Time Series Forecasting with Gaussian Processes Needs Priors}. In \bibinfo{booktitle}{\emph{Machine Learning and Knowledge Discovery in Databases. Applied Data Science Track - European Conference, {ECML} {PKDD} 2021}} \emph{(\bibinfo{series}{Lecture Notes in Computer Science}, Vol.~\bibinfo{volume}{12978})}. \bibinfo{publisher}{Springer}, \bibinfo{pages}{103--117}.
\newblock


\bibitem[Dai and Chen(2022)]%
        {DBLP:conf/iclr/DaiC22}
\bibfield{author}{\bibinfo{person}{Enyan Dai} {and} \bibinfo{person}{Jie Chen}.} \bibinfo{year}{2022}\natexlab{}.
\newblock \showarticletitle{Graph-Augmented Normalizing Flows for Anomaly Detection of Multiple Time Series}. In \bibinfo{booktitle}{\emph{The Tenth International Conference on Learning Representations, {ICLR} 2022}}.
\newblock


\bibitem[Dao and Gu(2024)]%
        {DBLP:conf/icml/DaoG24}
\bibfield{author}{\bibinfo{person}{Tri Dao} {and} \bibinfo{person}{Albert Gu}.} \bibinfo{year}{2024}\natexlab{}.
\newblock \showarticletitle{Transformers are SSMs: Generalized Models and Efficient Algorithms Through Structured State Space Duality}. In \bibinfo{booktitle}{\emph{Forty-first International Conference on Machine Learning, {ICML} 2024,}}.
\newblock


\bibitem[Desai et~al\mbox{.}(2021)]%
        {DBLP:journals/corr/abs-2111-08095}
\bibfield{author}{\bibinfo{person}{Abhyuday Desai}, \bibinfo{person}{Cynthia Freeman}, \bibinfo{person}{Zuhui Wang}, {and} \bibinfo{person}{Ian Beaver}.} \bibinfo{year}{2021}\natexlab{}.
\newblock \showarticletitle{TimeVAE: {A} Variational Auto-Encoder for Multivariate Time Series Generation}.
\newblock \bibinfo{journal}{\emph{CoRR}}  \bibinfo{volume}{abs/2111.08095} (\bibinfo{year}{2021}).
\newblock


\bibitem[Dinh et~al\mbox{.}(2017)]%
        {DBLP:conf/iclr/DinhSB17}
\bibfield{author}{\bibinfo{person}{Laurent Dinh}, \bibinfo{person}{Jascha Sohl{-}Dickstein}, {and} \bibinfo{person}{Samy Bengio}.} \bibinfo{year}{2017}\natexlab{}.
\newblock \showarticletitle{Density estimation using Real {NVP}}. In \bibinfo{booktitle}{\emph{5th International Conference on Learning Representations, {ICLR} 2017, Conference Track Proceedings}}. \bibinfo{publisher}{OpenReview.net}.
\newblock


\bibitem[Du et~al\mbox{.}(2023)]%
        {DBLP:journals/eswa/DuCL23}
\bibfield{author}{\bibinfo{person}{Wenjie Du}, \bibinfo{person}{David C{\^{o}}t{\'{e}}}, {and} \bibinfo{person}{Yan Liu}.} \bibinfo{year}{2023}\natexlab{}.
\newblock \showarticletitle{{SAITS:} Self-attention-based imputation for time series}.
\newblock \bibinfo{journal}{\emph{Expert Syst. Appl.}}  \bibinfo{volume}{219} (\bibinfo{year}{2023}), \bibinfo{pages}{119619}.
\newblock


\bibitem[Fang et~al\mbox{.}(2024)]%
        {DBLP:conf/icml/FangWLZ024}
\bibfield{author}{\bibinfo{person}{Shikai Fang}, \bibinfo{person}{Qingsong Wen}, \bibinfo{person}{Yingtao Luo}, \bibinfo{person}{Shandian Zhe}, {and} \bibinfo{person}{Liang Sun}.} \bibinfo{year}{2024}\natexlab{}.
\newblock \showarticletitle{BayOTIDE: Bayesian Online Multivariate Time Series Imputation with Functional Decomposition}. In \bibinfo{booktitle}{\emph{Forty-first International Conference on Machine Learning, {ICML} 2024}}.
\newblock


\bibitem[Fortuin et~al\mbox{.}(2020)]%
        {DBLP:conf/aistats/FortuinBRM20}
\bibfield{author}{\bibinfo{person}{Vincent Fortuin}, \bibinfo{person}{Dmitry Baranchuk}, \bibinfo{person}{Gunnar R{\"{a}}tsch}, {and} \bibinfo{person}{Stephan Mandt}.} \bibinfo{year}{2020}\natexlab{}.
\newblock \bibinfo{title}{{GP-VAE:} Deep Probabilistic Time Series Imputation}.
\newblock \bibinfo{numpages}{1651--1661}~pages.
\newblock


\bibitem[Fuller and Hasza(1981)]%
        {fuller1981properties}
\bibfield{author}{\bibinfo{person}{Wayne~A Fuller} {and} \bibinfo{person}{David~P Hasza}.} \bibinfo{year}{1981}\natexlab{}.
\newblock \showarticletitle{Properties of predictors for autoregressive time series}.
\newblock \bibinfo{journal}{\emph{J. Amer. Statist. Assoc.}} \bibinfo{volume}{76}, \bibinfo{number}{373} (\bibinfo{year}{1981}), \bibinfo{pages}{155--161}.
\newblock


\bibitem[Goodfellow et~al\mbox{.}(2014)]%
        {DBLP:conf/nips/GoodfellowPMXWOCB14}
\bibfield{author}{\bibinfo{person}{Ian~J. Goodfellow}, \bibinfo{person}{Jean Pouget{-}Abadie}, \bibinfo{person}{Mehdi Mirza}, \bibinfo{person}{Bing Xu}, \bibinfo{person}{David Warde{-}Farley}, \bibinfo{person}{Sherjil Ozair}, \bibinfo{person}{Aaron~C. Courville}, {and} \bibinfo{person}{Yoshua Bengio}.} \bibinfo{year}{2014}\natexlab{}.
\newblock \showarticletitle{Generative Adversarial Nets}. In \bibinfo{booktitle}{\emph{Advances in Neural Information Processing Systems 27: Annual Conference on Neural Information Processing Systems 2014}}. \bibinfo{pages}{2672--2680}.
\newblock


\bibitem[Gu and Dao(2023)]%
        {DBLP:journals/corr/abs-2312-00752}
\bibfield{author}{\bibinfo{person}{Albert Gu} {and} \bibinfo{person}{Tri Dao}.} \bibinfo{year}{2023}\natexlab{}.
\newblock \bibinfo{title}{Mamba: Linear-Time Sequence Modeling with Selective State Spaces}.
\newblock


\bibitem[Gu et~al\mbox{.}(2020)]%
        {DBLP:conf/nips/GuDERR20}
\bibfield{author}{\bibinfo{person}{Albert Gu}, \bibinfo{person}{Tri Dao}, \bibinfo{person}{Stefano Ermon}, \bibinfo{person}{Atri Rudra}, {and} \bibinfo{person}{Christopher R{\'{e}}}.} \bibinfo{year}{2020}\natexlab{}.
\newblock \showarticletitle{HiPPO: Recurrent Memory with Optimal Polynomial Projections}. In \bibinfo{booktitle}{\emph{Advances in Neural Information Processing Systems 33: Annual Conference on Neural Information Processing Systems 2020, NeurIPS 2020}}.
\newblock


\bibitem[Gu et~al\mbox{.}(2022)]%
        {DBLP:conf/iclr/GuGR22}
\bibfield{author}{\bibinfo{person}{Albert Gu}, \bibinfo{person}{Karan Goel}, {and} \bibinfo{person}{Christopher R{\'{e}}}.} \bibinfo{year}{2022}\natexlab{}.
\newblock \showarticletitle{Efficiently Modeling Long Sequences with Structured State Spaces}. In \bibinfo{booktitle}{\emph{The Tenth International Conference on Learning Representations, {ICLR} 2022, Virtual Event, April 25-29, 2022}}. \bibinfo{publisher}{OpenReview.net}.
\newblock


\bibitem[Guo et~al\mbox{.}(2014)]%
        {DBLP:journals/sigmod/GuoJ014}
\bibfield{author}{\bibinfo{person}{Chenjuan Guo}, \bibinfo{person}{Christian~S. Jensen}, {and} \bibinfo{person}{Bin Yang}.} \bibinfo{year}{2014}\natexlab{}.
\newblock \showarticletitle{Towards Total Traffic Awareness}.
\newblock \bibinfo{journal}{\emph{{SIGMOD} Record}} \bibinfo{volume}{43}, \bibinfo{number}{3} (\bibinfo{year}{2014}), \bibinfo{pages}{18--23}.
\newblock


\bibitem[Ho et~al\mbox{.}(2020)]%
        {DBLP:conf/nips/HoJA20}
\bibfield{author}{\bibinfo{person}{Jonathan Ho}, \bibinfo{person}{Ajay Jain}, {and} \bibinfo{person}{Pieter Abbeel}.} \bibinfo{year}{2020}\natexlab{}.
\newblock \showarticletitle{Denoising Diffusion Probabilistic Models}. In \bibinfo{booktitle}{\emph{Advances in Neural Information Processing Systems 33: Annual Conference on Neural Information Processing Systems 2020, NeurIPS 2020}}.
\newblock


\bibitem[Hochreiter and Schmidhuber(1997)]%
        {DBLP:journals/neco/HochreiterS97}
\bibfield{author}{\bibinfo{person}{Sepp Hochreiter} {and} \bibinfo{person}{J{\"{u}}rgen Schmidhuber}.} \bibinfo{year}{1997}\natexlab{}.
\newblock \showarticletitle{Long Short-Term Memory}.
\newblock \bibinfo{journal}{\emph{Neural Comput.}} \bibinfo{volume}{9}, \bibinfo{number}{8} (\bibinfo{year}{1997}), \bibinfo{pages}{1735--1780}.
\newblock


\bibitem[Hu et~al\mbox{.}(2024)]%
        {DBLP:journals/corr/abs-2405-16312}
\bibfield{author}{\bibinfo{person}{Jiaxi Hu}, \bibinfo{person}{Disen Lan}, \bibinfo{person}{Ziyu Zhou}, \bibinfo{person}{Qingsong Wen}, {and} \bibinfo{person}{Yuxuan Liang}.} \bibinfo{year}{2024}\natexlab{}.
\newblock \showarticletitle{Time-SSM: Simplifying and Unifying State Space Models for Time Series Forecasting}.
\newblock \bibinfo{journal}{\emph{CoRR}}  \bibinfo{volume}{abs/2405.16312} (\bibinfo{year}{2024}).
\newblock


\bibitem[Hu et~al\mbox{.}(2018)]%
        {DBLP:conf/cvpr/HuSS18}
\bibfield{author}{\bibinfo{person}{Jie Hu}, \bibinfo{person}{Li Shen}, {and} \bibinfo{person}{Gang Sun}.} \bibinfo{year}{2018}\natexlab{}.
\newblock \showarticletitle{Squeeze-and-Excitation Networks}. In \bibinfo{booktitle}{\emph{2018 {IEEE} Conference on Computer Vision and Pattern Recognition, {CVPR} 2018}}. \bibinfo{publisher}{Computer Vision Foundation / {IEEE} Computer Society}, \bibinfo{pages}{7132--7141}.
\newblock


\bibitem[Huang(2023)]%
        {DBLP:journals/corr/abs-2310-14341}
\bibfield{author}{\bibinfo{person}{Yexin Huang}.} \bibinfo{year}{2023}\natexlab{}.
\newblock \showarticletitle{Pyramidal Hidden Markov Model For Multivariate Time Series Forecasting}.
\newblock \bibinfo{journal}{\emph{CoRR}}  \bibinfo{volume}{abs/2310.14341} (\bibinfo{year}{2023}).
\newblock


\bibitem[Karevan and Suykens(2020)]%
        {DBLP:journals/nn/KarevanS20}
\bibfield{author}{\bibinfo{person}{Zahra Karevan} {and} \bibinfo{person}{Johan A.~K. Suykens}.} \bibinfo{year}{2020}\natexlab{}.
\newblock \showarticletitle{Transductive {LSTM} for time-series prediction: An application to weather forecasting}.
\newblock \bibinfo{journal}{\emph{Neural Networks}}  \bibinfo{volume}{125} (\bibinfo{year}{2020}), \bibinfo{pages}{1--9}.
\newblock


\bibitem[Kim et~al\mbox{.}(2023)]%
        {DBLP:conf/icml/KimKYLL023}
\bibfield{author}{\bibinfo{person}{Seunghyun Kim}, \bibinfo{person}{Hyunsu Kim}, \bibinfo{person}{Eunggu Yun}, \bibinfo{person}{Hwangrae Lee}, \bibinfo{person}{Jaehun Lee}, {and} \bibinfo{person}{Juho Lee}.} \bibinfo{year}{2023}\natexlab{}.
\newblock \bibinfo{title}{Probabilistic Imputation for Time-series Classification with Missing Data}.
\newblock \bibinfo{numpages}{16654--16667}~pages.
\newblock


\bibitem[Kingma and Ba(2015)]%
        {DBLP:journals/corr/KingmaB14}
\bibfield{author}{\bibinfo{person}{Diederik~P. Kingma} {and} \bibinfo{person}{Jimmy Ba}.} \bibinfo{year}{2015}\natexlab{}.
\newblock \showarticletitle{Adam: {A} Method for Stochastic Optimization}. In \bibinfo{booktitle}{\emph{3rd International Conference on Learning Representations, {ICLR} 2015,}}.
\newblock


\bibitem[Kingma and Welling(2014)]%
        {DBLP:journals/corr/KingmaW13}
\bibfield{author}{\bibinfo{person}{Diederik~P. Kingma} {and} \bibinfo{person}{Max Welling}.} \bibinfo{year}{2014}\natexlab{}.
\newblock \showarticletitle{Auto-Encoding Variational Bayes}. In \bibinfo{booktitle}{\emph{2nd International Conference on Learning Representations, {ICLR} 2014,}}.
\newblock


\bibitem[Kobyzev et~al\mbox{.}(2021)]%
        {DBLP:journals/pami/KobyzevPB21}
\bibfield{author}{\bibinfo{person}{Ivan Kobyzev}, \bibinfo{person}{Simon J.~D. Prince}, {and} \bibinfo{person}{Marcus~A. Brubaker}.} \bibinfo{year}{2021}\natexlab{}.
\newblock \showarticletitle{Normalizing Flows: An Introduction and Review of Current Methods}.
\newblock \bibinfo{journal}{\emph{{IEEE} Trans. Pattern Anal. Mach. Intell.}} \bibinfo{volume}{43}, \bibinfo{number}{11} (\bibinfo{year}{2021}), \bibinfo{pages}{3964--3979}.
\newblock


\bibitem[Kocadagli and Asikgil(2014)]%
        {DBLP:journals/eswa/KocadagliA14}
\bibfield{author}{\bibinfo{person}{Ozan Kocadagli} {and} \bibinfo{person}{Baris Asikgil}.} \bibinfo{year}{2014}\natexlab{}.
\newblock \showarticletitle{Nonlinear time series forecasting with Bayesian neural networks}.
\newblock \bibinfo{journal}{\emph{Expert Syst. Appl.}} \bibinfo{volume}{41}, \bibinfo{number}{15} (\bibinfo{year}{2014}), \bibinfo{pages}{6596--6610}.
\newblock


\bibitem[Lea et~al\mbox{.}(2017)]%
        {DBLP:conf/cvpr/LeaFVRH17}
\bibfield{author}{\bibinfo{person}{Colin Lea}, \bibinfo{person}{Michael~D. Flynn}, \bibinfo{person}{Ren{\'{e}} Vidal}, \bibinfo{person}{Austin Reiter}, {and} \bibinfo{person}{Gregory~D. Hager}.} \bibinfo{year}{2017}\natexlab{}.
\newblock \showarticletitle{Temporal Convolutional Networks for Action Segmentation and Detection}. In \bibinfo{booktitle}{\emph{2017 {IEEE} Conference on Computer Vision and Pattern Recognition, {CVPR} 2017}}. \bibinfo{publisher}{{IEEE} Computer Society}, \bibinfo{pages}{1003--1012}.
\newblock


\bibitem[Li et~al\mbox{.}(2021a)]%
        {DBLP:conf/cvpr/LiW0021}
\bibfield{author}{\bibinfo{person}{Jing Li}, \bibinfo{person}{Botong Wu}, \bibinfo{person}{Xinwei Sun}, {and} \bibinfo{person}{Yizhou Wang}.} \bibinfo{year}{2021}\natexlab{a}.
\newblock \showarticletitle{Causal Hidden Markov Model for Time Series Disease Forecasting}. In \bibinfo{booktitle}{\emph{{IEEE} Conference on Computer Vision and Pattern Recognition, {CVPR} 2021}}. \bibinfo{publisher}{Computer Vision Foundation / {IEEE}}, \bibinfo{pages}{12105--12114}.
\newblock


\bibitem[Li et~al\mbox{.}(2021b)]%
        {DBLP:journals/tnn/LiYWJ21}
\bibfield{author}{\bibinfo{person}{Longyuan Li}, \bibinfo{person}{Junchi Yan}, \bibinfo{person}{Haiyang Wang}, {and} \bibinfo{person}{Yaohui Jin}.} \bibinfo{year}{2021}\natexlab{b}.
\newblock \showarticletitle{Anomaly Detection of Time Series With Smoothness-Inducing Sequential Variational Auto-Encoder}.
\newblock \bibinfo{journal}{\emph{{IEEE} Trans. Neural Networks Learn. Syst.}} \bibinfo{volume}{32}, \bibinfo{number}{3} (\bibinfo{year}{2021}), \bibinfo{pages}{1177--1191}.
\newblock


\bibitem[Li et~al\mbox{.}(2025)]%
        {li2024TSMF-Bench}
\bibfield{author}{\bibinfo{person}{Zhe Li}, \bibinfo{person}{Xiangfei Qiu}, \bibinfo{person}{Peng Chen}, \bibinfo{person}{Yihang Wang}, \bibinfo{person}{Hanyin Cheng}, \bibinfo{person}{Yang Shu}, \bibinfo{person}{Jilin Hu}, \bibinfo{person}{Chenjuan Guo}, \bibinfo{person}{Aoying Zhou}, \bibinfo{person}{Christian~S. Jensen}, {and} \bibinfo{person}{Bin Yang}.} \bibinfo{year}{2025}\natexlab{}.
\newblock \showarticletitle{TSMF-Bench: Comprehensive and Unified Benchmarking of Foundation Models for Time Series Forecasting}. In \bibinfo{booktitle}{\emph{SIGKDD}}.
\newblock


\bibitem[Liu et~al\mbox{.}(2023)]%
        {DBLP:conf/iclr/LiuLCCJ23}
\bibfield{author}{\bibinfo{person}{Shuai Liu}, \bibinfo{person}{Xiucheng Li}, \bibinfo{person}{Gao Cong}, \bibinfo{person}{Yile Chen}, {and} \bibinfo{person}{Yue Jiang}.} \bibinfo{year}{2023}\natexlab{}.
\newblock \showarticletitle{Multivariate Time-series Imputation with Disentangled Temporal Representations}. In \bibinfo{booktitle}{\emph{The Eleventh International Conference on Learning Representations, {ICLR} 2023}}. \bibinfo{publisher}{OpenReview.net}.
\newblock


\bibitem[Liu et~al\mbox{.}(2025)]%
        {DBLP:conf/iclr/LiuSGY25}
\bibfield{author}{\bibinfo{person}{Sicong Liu}, \bibinfo{person}{Yang Shu}, \bibinfo{person}{Chenjuan Guo}, {and} \bibinfo{person}{Bin Yang}.} \bibinfo{year}{2025}\natexlab{}.
\newblock \showarticletitle{Learning Generalizable Skills from Offline Multi-Task Data for Multi-Agent Cooperation}. In \bibinfo{booktitle}{\emph{The Thirteenth International Conference on Learning Representations, {ICLR}}}. \bibinfo{publisher}{OpenReview.net}.
\newblock


\bibitem[Luo et~al\mbox{.}(2018)]%
        {DBLP:conf/nips/LuoCZXY18}
\bibfield{author}{\bibinfo{person}{Yonghong Luo}, \bibinfo{person}{Xiangrui Cai}, \bibinfo{person}{Ying Zhang}, \bibinfo{person}{Jun Xu}, {and} \bibinfo{person}{Xiaojie Yuan}.} \bibinfo{year}{2018}\natexlab{}.
\newblock \bibinfo{title}{Multivariate Time Series Imputation with Generative Adversarial Networks}.
\newblock \bibinfo{numpages}{1603--1614}~pages.
\newblock


\bibitem[Ma et~al\mbox{.}(2024)]%
        {DBLP:journals/corr/abs-2411-02941}
\bibfield{author}{\bibinfo{person}{Haoyu Ma}, \bibinfo{person}{Yushu Chen}, \bibinfo{person}{Wenlai Zhao}, \bibinfo{person}{Jinzhe Yang}, \bibinfo{person}{Yingsheng Ji}, \bibinfo{person}{Xinghua Xu}, \bibinfo{person}{Xiaozhu Liu}, \bibinfo{person}{Hao Jing}, \bibinfo{person}{Shengzhuo Liu}, {and} \bibinfo{person}{Guangwen Yang}.} \bibinfo{year}{2024}\natexlab{}.
\newblock \showarticletitle{A Mamba Foundation Model for Time Series Forecasting}.
\newblock \bibinfo{journal}{\emph{CoRR}}  \bibinfo{volume}{abs/2411.02941} (\bibinfo{year}{2024}).
\newblock


\bibitem[Marisca et~al\mbox{.}(2022)]%
        {DBLP:conf/nips/MariscaCA22}
\bibfield{author}{\bibinfo{person}{Ivan Marisca}, \bibinfo{person}{Andrea Cini}, {and} \bibinfo{person}{Cesare Alippi}.} \bibinfo{year}{2022}\natexlab{}.
\newblock \showarticletitle{Learning to Reconstruct Missing Data from Spatiotemporal Graphs with Sparse Observations}. In \bibinfo{booktitle}{\emph{Advances in Neural Information Processing Systems 35: Annual Conference on Neural Information Processing Systems 2022, NeurIPS 2022}}, \bibfield{editor}{\bibinfo{person}{Sanmi Koyejo}, \bibinfo{person}{S.~Mohamed}, \bibinfo{person}{A.~Agarwal}, \bibinfo{person}{Danielle Belgrave}, \bibinfo{person}{K.~Cho}, {and} \bibinfo{person}{A.~Oh}} (Eds.).
\newblock


\bibitem[McGovern et~al\mbox{.}(2011)]%
        {DBLP:journals/datamine/McGovernRBD11}
\bibfield{author}{\bibinfo{person}{Amy McGovern}, \bibinfo{person}{Derek~H. Rosendahl}, \bibinfo{person}{Rodger~A. Brown}, {and} \bibinfo{person}{Kelvin Droegemeier}.} \bibinfo{year}{2011}\natexlab{}.
\newblock \showarticletitle{Identifying predictive multi-dimensional time series motifs: an application to severe weather prediction}.
\newblock \bibinfo{journal}{\emph{Data Min. Knowl. Discov.}} \bibinfo{volume}{22}, \bibinfo{number}{1-2} (\bibinfo{year}{2011}), \bibinfo{pages}{232--258}.
\newblock


\bibitem[Meek et~al\mbox{.}(2002)]%
        {meek2002autoregressive}
\bibfield{author}{\bibinfo{person}{Christopher Meek}, \bibinfo{person}{David~Maxwell Chickering}, {and} \bibinfo{person}{David Heckerman}.} \bibinfo{year}{2002}\natexlab{}.
\newblock \showarticletitle{Autoregressive tree models for time-series analysis}. In \bibinfo{booktitle}{\emph{Proceedings of the 2002 SIAM International Conference on Data Mining}}. SIAM, \bibinfo{pages}{229--244}.
\newblock


\bibitem[Miao et~al\mbox{.}(2024)]%
        {DBLP:conf/icde/00010GY0HXJ24}
\bibfield{author}{\bibinfo{person}{Hao Miao}, \bibinfo{person}{Yan Zhao}, \bibinfo{person}{Chenjuan Guo}, \bibinfo{person}{Bin Yang}, \bibinfo{person}{Kai Zheng}, \bibinfo{person}{Feiteng Huang}, \bibinfo{person}{Jiandong Xie}, {and} \bibinfo{person}{Christian~S. Jensen}.} \bibinfo{year}{2024}\natexlab{}.
\newblock \showarticletitle{A Unified Replay-Based Continuous Learning Framework for Spatio-Temporal Prediction on Streaming Data}. In \bibinfo{booktitle}{\emph{{ICDE}}}. \bibinfo{pages}{1050--1062}.
\newblock


\bibitem[Miao et~al\mbox{.}(2021)]%
        {DBLP:conf/aaai/MiaoWWGMY21}
\bibfield{author}{\bibinfo{person}{Xiaoye Miao}, \bibinfo{person}{Yangyang Wu}, \bibinfo{person}{Jun Wang}, \bibinfo{person}{Yunjun Gao}, \bibinfo{person}{Xudong Mao}, {and} \bibinfo{person}{Jianwei Yin}.} \bibinfo{year}{2021}\natexlab{}.
\newblock \showarticletitle{Generative Semi-supervised Learning for Multivariate Time Series Imputation}. In \bibinfo{booktitle}{\emph{Thirty-Fifth {AAAI} Conference on Artificial Intelligence, {AAAI} 2021}}. \bibinfo{publisher}{{AAAI} Press}, \bibinfo{pages}{8983--8991}.
\newblock


\bibitem[Morid et~al\mbox{.}(2023)]%
        {DBLP:journals/tmis/MoridSD23}
\bibfield{author}{\bibinfo{person}{Mohammad~Amin Morid}, \bibinfo{person}{Olivia R.~Liu Sheng}, {and} \bibinfo{person}{Joseph Dunbar}.} \bibinfo{year}{2023}\natexlab{}.
\newblock \bibinfo{title}{Time Series Prediction Using Deep Learning Methods in Healthcare}.
\newblock \bibinfo{numpages}{2:1--2:29}~pages.
\newblock


\bibitem[Morrill et~al\mbox{.}(2021)]%
        {DBLP:conf/icml/MorrillSKF21}
\bibfield{author}{\bibinfo{person}{James Morrill}, \bibinfo{person}{Cristopher Salvi}, \bibinfo{person}{Patrick Kidger}, {and} \bibinfo{person}{James Foster}.} \bibinfo{year}{2021}\natexlab{}.
\newblock \showarticletitle{Neural Rough Differential Equations for Long Time Series}. In \bibinfo{booktitle}{\emph{Proceedings of the 38th International Conference on Machine Learning,{ICML} 2021}} \emph{(\bibinfo{series}{Proceedings of Machine Learning Research}, Vol.~\bibinfo{volume}{139})}. \bibinfo{publisher}{{PMLR}}, \bibinfo{pages}{7829--7838}.
\newblock


\bibitem[Mulyadi et~al\mbox{.}(2022)]%
        {DBLP:journals/tcyb/MulyadiJS22}
\bibfield{author}{\bibinfo{person}{Ahmad~Wisnu Mulyadi}, \bibinfo{person}{Eunji Jun}, {and} \bibinfo{person}{Heung{-}Il Suk}.} \bibinfo{year}{2022}\natexlab{}.
\newblock \showarticletitle{Uncertainty-Aware Variational-Recurrent Imputation Network for Clinical Time Series}.
\newblock \bibinfo{journal}{\emph{{IEEE} Trans. Cybern.}} \bibinfo{volume}{52}, \bibinfo{number}{9} (\bibinfo{year}{2022}), \bibinfo{pages}{9684--9694}.
\newblock


\bibitem[Nelson(1998)]%
        {nelson1998time}
\bibfield{author}{\bibinfo{person}{Brian~K Nelson}.} \bibinfo{year}{1998}\natexlab{}.
\newblock \showarticletitle{Time series analysis using autoregressive integrated moving average (ARIMA) models}.
\newblock \bibinfo{journal}{\emph{Academic emergency medicine}} \bibinfo{volume}{5}, \bibinfo{number}{7} (\bibinfo{year}{1998}), \bibinfo{pages}{739--744}.
\newblock


\bibitem[Newbold(1983)]%
        {newbold1983arima}
\bibfield{author}{\bibinfo{person}{Paul Newbold}.} \bibinfo{year}{1983}\natexlab{}.
\newblock \showarticletitle{ARIMA model building and the time series analysis approach to forecasting}.
\newblock \bibinfo{journal}{\emph{Journal of forecasting}} \bibinfo{volume}{2}, \bibinfo{number}{1} (\bibinfo{year}{1983}), \bibinfo{pages}{23--35}.
\newblock


\bibitem[Nie et~al\mbox{.}(2024)]%
        {DBLP:conf/kdd/NieQMMS24}
\bibfield{author}{\bibinfo{person}{Tong Nie}, \bibinfo{person}{Guoyang Qin}, \bibinfo{person}{Wei Ma}, \bibinfo{person}{Yuewen Mei}, {and} \bibinfo{person}{Jian Sun}.} \bibinfo{year}{2024}\natexlab{}.
\newblock \showarticletitle{ImputeFormer: Low Rankness-Induced Transformers for Generalizable Spatiotemporal Imputation}. In \bibinfo{booktitle}{\emph{Proceedings of the 30th {ACM} {SIGKDD} Conference on Knowledge Discovery and Data Mining, {KDD} 2024}}. \bibinfo{publisher}{{ACM}}, \bibinfo{pages}{2260--2271}.
\newblock


\bibitem[Nie et~al\mbox{.}(2023)]%
        {DBLP:conf/iclr/NieNSK23}
\bibfield{author}{\bibinfo{person}{Yuqi Nie}, \bibinfo{person}{Nam~H. Nguyen}, \bibinfo{person}{Phanwadee Sinthong}, {and} \bibinfo{person}{Jayant Kalagnanam}.} \bibinfo{year}{2023}\natexlab{}.
\newblock \showarticletitle{A Time Series is Worth 64 Words: Long-term Forecasting with Transformers}. In \bibinfo{booktitle}{\emph{The Eleventh International Conference on Learning Representations, {ICLR} 2023}}.
\newblock


\bibitem[Ouyang et~al\mbox{.}(2025)]%
        {DBLP:journals/corr/abs-2503-04252}
\bibfield{author}{\bibinfo{person}{Biao Ouyang}, \bibinfo{person}{Yingying Zhang}, \bibinfo{person}{Hanyin Cheng}, \bibinfo{person}{Yang Shu}, \bibinfo{person}{Chenjuan Guo}, \bibinfo{person}{Bin Yang}, \bibinfo{person}{Qingsong Wen}, \bibinfo{person}{Lunting Fan}, {and} \bibinfo{person}{Christian~S. Jensen}.} \bibinfo{year}{2025}\natexlab{}.
\newblock \showarticletitle{RCRank: Multimodal Ranking of Root Causes of Slow Queries in Cloud Database Systems}.
\newblock \bibinfo{journal}{\emph{CoRR}}  \bibinfo{volume}{abs/2503.04252} (\bibinfo{year}{2025}).
\newblock


\bibitem[Owusu et~al\mbox{.}(2023)]%
        {DBLP:conf/icdm/OwusuTPBW23}
\bibfield{author}{\bibinfo{person}{Patrick~Asante Owusu}, \bibinfo{person}{Etienne~Gael Tajeuna}, \bibinfo{person}{Jean{-}Marc Patenaude}, \bibinfo{person}{Armelle Brun}, {and} \bibinfo{person}{Shengrui Wang}.} \bibinfo{year}{2023}\natexlab{}.
\newblock \bibinfo{title}{Rethinking Temporal Dependencies in Multiple Time Series: {A} Use Case in Financial Data}.
\newblock \bibinfo{numpages}{1247--1252}~pages.
\newblock


\bibitem[Park et~al\mbox{.}(2023)]%
        {DBLP:journals/nca/ParkMAFPVSA23}
\bibfield{author}{\bibinfo{person}{Jangho Park}, \bibinfo{person}{Juliane M{\"{u}}ller}, \bibinfo{person}{Bhavna Arora}, \bibinfo{person}{Boris Faybishenko}, \bibinfo{person}{Gilberto~Zonta Pastorello}, \bibinfo{person}{Charuleka Varadharajan}, \bibinfo{person}{Reetik Sahu}, {and} \bibinfo{person}{Deborah~A. Agarwal}.} \bibinfo{year}{2023}\natexlab{}.
\newblock \showarticletitle{Long-term missing value imputation for time series data using deep neural networks}.
\newblock \bibinfo{journal}{\emph{Neural Comput. Appl.}} \bibinfo{volume}{35}, \bibinfo{number}{12} (\bibinfo{year}{2023}), \bibinfo{pages}{9071--9091}.
\newblock


\bibitem[Paszke et~al\mbox{.}(2019)]%
        {paszke2019pytorch}
\bibfield{author}{\bibinfo{person}{Adam Paszke}, \bibinfo{person}{Sam Gross}, \bibinfo{person}{Francisco Massa}, \bibinfo{person}{Adam Lerer}, \bibinfo{person}{James Bradbury}, \bibinfo{person}{Gregory Chanan}, \bibinfo{person}{Trevor Killeen}, \bibinfo{person}{Zeming Lin}, \bibinfo{person}{Natalia Gimelshein}, \bibinfo{person}{Luca Antiga}, {et~al\mbox{.}}} \bibinfo{year}{2019}\natexlab{}.
\newblock \bibinfo{title}{Pytorch: An imperative style, high-performance deep learning library}.
\newblock


\bibitem[Poyraz and Marttinen(2023)]%
        {DBLP:conf/ml4h/PoyrazM23}
\bibfield{author}{\bibinfo{person}{Onur Poyraz} {and} \bibinfo{person}{Pekka Marttinen}.} \bibinfo{year}{2023}\natexlab{}.
\newblock \bibinfo{title}{Mixture of Coupled HMMs for Robust Modeling of Multivariate Healthcare Time Series}.
\newblock \bibinfo{numpages}{461--479}~pages.
\newblock


\bibitem[Qiu et~al\mbox{.}(2024)]%
        {qiu2024tfb}
\bibfield{author}{\bibinfo{person}{Xiangfei Qiu}, \bibinfo{person}{Jilin Hu}, \bibinfo{person}{Lekui Zhou}, \bibinfo{person}{Xingjian Wu}, \bibinfo{person}{Junyang Du}, \bibinfo{person}{Buang Zhang}, \bibinfo{person}{Chenjuan Guo}, \bibinfo{person}{Aoying Zhou}, \bibinfo{person}{Christian~S. Jensen}, \bibinfo{person}{Zhenli Sheng}, {and} \bibinfo{person}{Bin Yang}.} \bibinfo{year}{2024}\natexlab{}.
\newblock \showarticletitle{TFB: Towards Comprehensive and Fair Benchmarking of Time Series Forecasting Methods}. In \bibinfo{booktitle}{\emph{Proc. {VLDB} Endow.}} \bibinfo{pages}{2363--2377}.
\newblock


\bibitem[Qiu et~al\mbox{.}(2025a)]%
        {qiu2025tab}
\bibfield{author}{\bibinfo{person}{Xiangfei Qiu}, \bibinfo{person}{Zhe Li}, \bibinfo{person}{Wanghui Qiu}, \bibinfo{person}{Shiyan Hu}, \bibinfo{person}{Lekui Zhou}, \bibinfo{person}{Xingjian Wu}, \bibinfo{person}{Zhengyu Li}, \bibinfo{person}{Chenjuan Guo}, \bibinfo{person}{Aoying Zhou}, \bibinfo{person}{Zhenli Sheng}, \bibinfo{person}{Jilin Hu}, \bibinfo{person}{Christian~S. Jensen}, {and} \bibinfo{person}{Bin Yang}.} \bibinfo{year}{2025}\natexlab{a}.
\newblock \showarticletitle{TAB: Unified Benchmarking of Time Series Anomaly Detection Methods}. In \bibinfo{booktitle}{\emph{Proc. {VLDB} Endow.}}
\newblock


\bibitem[Qiu et~al\mbox{.}(2025b)]%
        {qiu2025duet}
\bibfield{author}{\bibinfo{person}{Xiangfei Qiu}, \bibinfo{person}{Xingjian Wu}, \bibinfo{person}{Yan Lin}, \bibinfo{person}{Chenjuan Guo}, \bibinfo{person}{Jilin Hu}, {and} \bibinfo{person}{Bin Yang}.} \bibinfo{year}{2025}\natexlab{b}.
\newblock \showarticletitle{DUET: Dual Clustering Enhanced Multivariate Time Series Forecasting}. In \bibinfo{booktitle}{\emph{SIGKDD}}. \bibinfo{pages}{1185--1196}.
\newblock


\bibitem[Rasul et~al\mbox{.}(2021a)]%
        {DBLP:conf/icml/RasulSSV21}
\bibfield{author}{\bibinfo{person}{Kashif Rasul}, \bibinfo{person}{Calvin Seward}, \bibinfo{person}{Ingmar Schuster}, {and} \bibinfo{person}{Roland Vollgraf}.} \bibinfo{year}{2021}\natexlab{a}.
\newblock \showarticletitle{Autoregressive Denoising Diffusion Models for Multivariate Probabilistic Time Series Forecasting}. In \bibinfo{booktitle}{\emph{Proceedings of the 38th International Conference on Machine Learning, {ICML} 2021}} \emph{(\bibinfo{series}{Proceedings of Machine Learning Research}, Vol.~\bibinfo{volume}{139})}. \bibinfo{publisher}{{PMLR}}, \bibinfo{pages}{8857--8868}.
\newblock


\bibitem[Rasul et~al\mbox{.}(2021b)]%
        {DBLP:conf/iclr/RasulSSBV21}
\bibfield{author}{\bibinfo{person}{Kashif Rasul}, \bibinfo{person}{Abdul{-}Saboor Sheikh}, \bibinfo{person}{Ingmar Schuster}, \bibinfo{person}{Urs~M. Bergmann}, {and} \bibinfo{person}{Roland Vollgraf}.} \bibinfo{year}{2021}\natexlab{b}.
\newblock \showarticletitle{Multivariate Probabilistic Time Series Forecasting via Conditioned Normalizing Flows}. In \bibinfo{booktitle}{\emph{9th International Conference on Learning Representations, {ICLR} 2021}}. \bibinfo{publisher}{OpenReview.net}.
\newblock


\bibitem[Ronneberger et~al\mbox{.}(2015)]%
        {DBLP:conf/miccai/RonnebergerFB15}
\bibfield{author}{\bibinfo{person}{Olaf Ronneberger}, \bibinfo{person}{Philipp Fischer}, {and} \bibinfo{person}{Thomas Brox}.} \bibinfo{year}{2015}\natexlab{}.
\newblock \showarticletitle{U-Net: Convolutional Networks for Biomedical Image Segmentation}. In \bibinfo{booktitle}{\emph{Medical Image Computing and Computer-Assisted Intervention - {MICCAI}}} \emph{(\bibinfo{series}{Lecture Notes in Computer Science}, Vol.~\bibinfo{volume}{9351})}. \bibinfo{pages}{234--241}.
\newblock


\bibitem[Rubanova et~al\mbox{.}(2019)]%
        {DBLP:conf/nips/RubanovaCD19}
\bibfield{author}{\bibinfo{person}{Yulia Rubanova}, \bibinfo{person}{Tian~Qi Chen}, {and} \bibinfo{person}{David Duvenaud}.} \bibinfo{year}{2019}\natexlab{}.
\newblock \showarticletitle{Latent Ordinary Differential Equations for Irregularly-Sampled Time Series}. In \bibinfo{booktitle}{\emph{Advances in Neural Information Processing Systems 32: Annual Conference on Neural Information Processing Systems 2019, NeurIPS 2019}}. \bibinfo{pages}{5321--5331}.
\newblock


\bibitem[Seifner et~al\mbox{.}(2025)]%
        {DBLP:conf/iclr/SeifnerCKS25}
\bibfield{author}{\bibinfo{person}{Patrick Seifner}, \bibinfo{person}{Kostadin Cvejoski}, \bibinfo{person}{Antonia K{\"{o}}rner}, {and} \bibinfo{person}{Rams{\'{e}}s~J. S{\'{a}}nchez}.} \bibinfo{year}{2025}\natexlab{}.
\newblock \showarticletitle{Zero-shot Imputation with Foundation Inference Models for Dynamical Systems}. In \bibinfo{booktitle}{\emph{The Thirteenth International Conference on Learning Representations, {ICLR} 2025}}. \bibinfo{publisher}{OpenReview.net}.
\newblock


\bibitem[Shan et~al\mbox{.}(2023)]%
        {shan2023nrtsi}
\bibfield{author}{\bibinfo{person}{Siyuan Shan}, \bibinfo{person}{Yang Li}, {and} \bibinfo{person}{Junier~B Oliva}.} \bibinfo{year}{2023}\natexlab{}.
\newblock \bibinfo{title}{Nrtsi: Non-recurrent time series imputation}.
\newblock \bibinfo{numpages}{5}~pages.
\newblock


\bibitem[Shen et~al\mbox{.}(2024)]%
        {DBLP:conf/iclr/ShenCK24}
\bibfield{author}{\bibinfo{person}{Lifeng Shen}, \bibinfo{person}{Weiyu Chen}, {and} \bibinfo{person}{James~T. Kwok}.} \bibinfo{year}{2024}\natexlab{}.
\newblock \showarticletitle{Multi-Resolution Diffusion Models for Time Series Forecasting}. In \bibinfo{booktitle}{\emph{The Twelfth International Conference on Learning Representations, {ICLR} 2024, Vienna}}.
\newblock


\bibitem[Silva et~al\mbox{.}(2012)]%
        {silva2012predicting}
\bibfield{author}{\bibinfo{person}{Ikaro Silva}, \bibinfo{person}{George Moody}, \bibinfo{person}{Daniel~J Scott}, \bibinfo{person}{Leo~A Celi}, {and} \bibinfo{person}{Roger~G Mark}.} \bibinfo{year}{2012}\natexlab{}.
\newblock \showarticletitle{Predicting in-hospital mortality of icu patients: The physionet/computing in cardiology challenge 2012}. In \bibinfo{booktitle}{\emph{2012 computing in cardiology}}. IEEE, \bibinfo{pages}{245--248}.
\newblock


\bibitem[Suo et~al\mbox{.}(2020)]%
        {suo2020glima}
\bibfield{author}{\bibinfo{person}{Qiuling Suo}, \bibinfo{person}{Weida Zhong}, \bibinfo{person}{Guangxu Xun}, \bibinfo{person}{Jianhui Sun}, \bibinfo{person}{Changyou Chen}, {and} \bibinfo{person}{Aidong Zhang}.} \bibinfo{year}{2020}\natexlab{}.
\newblock \bibinfo{title}{GLIMA: Global and local time series imputation with multi-directional attention learning}.
\newblock \bibinfo{numpages}{798--807}~pages.
\newblock


\bibitem[Tashiro et~al\mbox{.}(2021)]%
        {DBLP:conf/nips/TashiroSSE21}
\bibfield{author}{\bibinfo{person}{Yusuke Tashiro}, \bibinfo{person}{Jiaming Song}, \bibinfo{person}{Yang Song}, {and} \bibinfo{person}{Stefano Ermon}.} \bibinfo{year}{2021}\natexlab{}.
\newblock \bibinfo{title}{{CSDI:} Conditional Score-based Diffusion Models for Probabilistic Time Series Imputation}.
\newblock \bibinfo{numpages}{24804--24816}~pages.
\newblock


\bibitem[Tian et~al\mbox{.}(2025)]%
        {Air-DualODE}
\bibfield{author}{\bibinfo{person}{Jindong Tian}, \bibinfo{person}{Yuxuan Liang}, \bibinfo{person}{Ronghui Xu}, \bibinfo{person}{Peng Chen}, \bibinfo{person}{Chenjuan Guo}, \bibinfo{person}{Aoying Zhou}, \bibinfo{person}{Lujia Pan}, \bibinfo{person}{Zhongwen Rao}, {and} \bibinfo{person}{Bin Yang}.} \bibinfo{year}{2025}\natexlab{}.
\newblock \showarticletitle{Air quality prediction with physics-guided dual neural odes in open systems}. In \bibinfo{booktitle}{\emph{The Thirteenth International Conference on Learning Representations}}.
\newblock


\bibitem[Tzelepi et~al\mbox{.}(2023)]%
        {DBLP:conf/icassp/TzelepiNT23}
\bibfield{author}{\bibinfo{person}{Maria Tzelepi}, \bibinfo{person}{Paraskevi Nousi}, {and} \bibinfo{person}{Anastasios Tefas}.} \bibinfo{year}{2023}\natexlab{}.
\newblock \bibinfo{title}{Improving Electric Load Demand Forecasting with Anchor-Based Forecasting Method}.
\newblock \bibinfo{numpages}{5}~pages.
\newblock


\bibitem[Vaswani et~al\mbox{.}(2017)]%
        {DBLP:conf/nips/VaswaniSPUJGKP17}
\bibfield{author}{\bibinfo{person}{Ashish Vaswani}, \bibinfo{person}{Noam Shazeer}, \bibinfo{person}{Niki Parmar}, \bibinfo{person}{Jakob Uszkoreit}, \bibinfo{person}{Llion Jones}, \bibinfo{person}{Aidan~N. Gomez}, \bibinfo{person}{Lukasz Kaiser}, {and} \bibinfo{person}{Illia Polosukhin}.} \bibinfo{year}{2017}\natexlab{}.
\newblock \showarticletitle{Attention is All you Need}. In \bibinfo{booktitle}{\emph{Advances in Neural Information Processing Systems 30: Annual Conference on Neural Information Processing Systems 2017}}. \bibinfo{pages}{5998--6008}.
\newblock


\bibitem[Wang et~al\mbox{.}(2025)]%
        {DBLP:journals/ijon/WangKFWYZWZ25}
\bibfield{author}{\bibinfo{person}{Zihan Wang}, \bibinfo{person}{Fanheng Kong}, \bibinfo{person}{Shi Feng}, \bibinfo{person}{Ming Wang}, \bibinfo{person}{Xiaocui Yang}, \bibinfo{person}{Han Zhao}, \bibinfo{person}{Daling Wang}, {and} \bibinfo{person}{Yifei Zhang}.} \bibinfo{year}{2025}\natexlab{}.
\newblock \showarticletitle{Is Mamba effective for time series forecasting?}
\newblock \bibinfo{journal}{\emph{Neurocomputing}}  \bibinfo{volume}{619} (\bibinfo{year}{2025}), \bibinfo{pages}{129178}.
\newblock


\bibitem[Wu et~al\mbox{.}(2025a)]%
        {wu2025k2vae}
\bibfield{author}{\bibinfo{person}{Xingjian Wu}, \bibinfo{person}{Xiangfei Qiu}, \bibinfo{person}{Hongfan Gao}, \bibinfo{person}{Jilin Hu}, \bibinfo{person}{Chenjuan Guo}, {and} \bibinfo{person}{Bin Yang}.} \bibinfo{year}{2025}\natexlab{a}.
\newblock \showarticletitle{K${}^2$VAE: A Koopman-Kalman Enhanced Variational AutoEncoder for Probabilistic Time Series Forecasting}. In \bibinfo{booktitle}{\emph{ICML}}.
\newblock


\bibitem[Wu et~al\mbox{.}(2025b)]%
        {wu2024catch}
\bibfield{author}{\bibinfo{person}{Xingjian Wu}, \bibinfo{person}{Xiangfei Qiu}, \bibinfo{person}{Zhengyu Li}, \bibinfo{person}{Yihang Wang}, \bibinfo{person}{Jilin Hu}, \bibinfo{person}{Chenjuan Guo}, \bibinfo{person}{Hui Xiong}, {and} \bibinfo{person}{Bin Yang}.} \bibinfo{year}{2025}\natexlab{b}.
\newblock \showarticletitle{CATCH: Channel-Aware multivariate Time Series Anomaly Detection via Frequency Patching}. In \bibinfo{booktitle}{\emph{ICLR}}.
\newblock


\bibitem[Wu et~al\mbox{.}(2023)]%
        {DBLP:journals/pacmmod/Wu0ZG0J23}
\bibfield{author}{\bibinfo{person}{Xinle Wu}, \bibinfo{person}{Dalin Zhang}, \bibinfo{person}{Miao Zhang}, \bibinfo{person}{Chenjuan Guo}, \bibinfo{person}{Bin Yang}, {and} \bibinfo{person}{Christian~S. Jensen}.} \bibinfo{year}{2023}\natexlab{}.
\newblock \showarticletitle{Auto{CTS}+: Joint Neural Architecture and Hyperparameter Search for Correlated Time Series Forecasting}.
\newblock \bibinfo{journal}{\emph{Proc. {ACM} Manag. Data}} \bibinfo{volume}{1}, \bibinfo{number}{1} (\bibinfo{year}{2023}), \bibinfo{pages}{97:1--97:26}.
\newblock


\bibitem[Xiang et~al\mbox{.}(2022)]%
        {DBLP:conf/cikm/XiangCSZL22}
\bibfield{author}{\bibinfo{person}{Sheng Xiang}, \bibinfo{person}{Dawei Cheng}, \bibinfo{person}{Chencheng Shang}, \bibinfo{person}{Ying Zhang}, {and} \bibinfo{person}{Yuqi Liang}.} \bibinfo{year}{2022}\natexlab{}.
\newblock \bibinfo{title}{Temporal and Heterogeneous Graph Neural Network for Financial Time Series Prediction}.
\newblock \bibinfo{numpages}{3584--3593}~pages.
\newblock


\bibitem[Xu et~al\mbox{.}(2025)]%
        {10.1145/3690624.3709209}
\bibfield{author}{\bibinfo{person}{Ronghui Xu}, \bibinfo{person}{Hanyin Cheng}, \bibinfo{person}{Chenjuan Guo}, \bibinfo{person}{Hongfan Gao}, \bibinfo{person}{Jilin Hu}, \bibinfo{person}{Sean~Bin Yang}, {and} \bibinfo{person}{Bin Yang}.} \bibinfo{year}{2025}\natexlab{}.
\newblock \showarticletitle{MM-Path: Multi-modal, Multi-granularity Path Representation Learning}. In \bibinfo{booktitle}{\emph{Proceedings of the 31st ACM SIGKDD Conference on Knowledge Discovery and Data Mining V.1}}. \bibinfo{pages}{1703–1714}.
\newblock


\bibitem[Xu et~al\mbox{.}(2023)]%
        {xu2023spatial}
\bibfield{author}{\bibinfo{person}{Ronghui Xu}, \bibinfo{person}{Weiming Huang}, \bibinfo{person}{Jun Zhao}, \bibinfo{person}{Meng Chen}, {and} \bibinfo{person}{Liqiang Nie}.} \bibinfo{year}{2023}\natexlab{}.
\newblock \showarticletitle{A Spatial and Adversarial Representation Learning Approach for Land Use Classification with POIs}.
\newblock \bibinfo{journal}{\emph{ACM Transactions on Intelligent Systems and Technology}} \bibinfo{volume}{14}, \bibinfo{number}{6} (\bibinfo{year}{2023}), \bibinfo{pages}{1--25}.
\newblock


\bibitem[Yan et~al\mbox{.}(2024)]%
        {DBLP:conf/icml/YanGHZX24}
\bibfield{author}{\bibinfo{person}{Tijin Yan}, \bibinfo{person}{Hengheng Gong}, \bibinfo{person}{Yongping He}, \bibinfo{person}{Yufeng Zhan}, {and} \bibinfo{person}{Yuanqing Xia}.} \bibinfo{year}{2024}\natexlab{}.
\newblock \showarticletitle{Probabilistic Time Series Modeling with Decomposable Denoising Diffusion Model}. In \bibinfo{booktitle}{\emph{Forty-first International Conference on Machine Learning, {ICML} 2024}}. \bibinfo{publisher}{OpenReview.net}.
\newblock


\bibitem[Yi et~al\mbox{.}(2016)]%
        {DBLP:conf/ijcai/YiZZL16}
\bibfield{author}{\bibinfo{person}{Xiuwen Yi}, \bibinfo{person}{Yu Zheng}, \bibinfo{person}{Junbo Zhang}, {and} \bibinfo{person}{Tianrui Li}.} \bibinfo{year}{2016}\natexlab{}.
\newblock \showarticletitle{{ST-MVL:} Filling Missing Values in Geo-Sensory Time Series Data}. In \bibinfo{booktitle}{\emph{Proceedings of the Twenty-Fifth International Joint Conference on Artificial Intelligence, {IJCAI} 2016}}. \bibinfo{publisher}{{IJCAI/AAAI} Press}, \bibinfo{pages}{2704--2710}.
\newblock


\bibitem[Yoon et~al\mbox{.}(2019)]%
        {DBLP:conf/nips/YoonJS19}
\bibfield{author}{\bibinfo{person}{Jinsung Yoon}, \bibinfo{person}{Daniel Jarrett}, {and} \bibinfo{person}{Mihaela van~der Schaar}.} \bibinfo{year}{2019}\natexlab{}.
\newblock \showarticletitle{Time-series Generative Adversarial Networks}. In \bibinfo{booktitle}{\emph{Advances in Neural Information Processing Systems 32: Annual Conference on Neural Information Processing Systems 2019, NeurIPS 2019}}. \bibinfo{pages}{5509--5519}.
\newblock


\bibitem[Yuan et~al\mbox{.}(2018)]%
        {DBLP:conf/huc/YuanXYJZ18}
\bibfield{author}{\bibinfo{person}{Hongwu Yuan}, \bibinfo{person}{Guoming Xu}, \bibinfo{person}{Zijian Yao}, \bibinfo{person}{Ji Jia}, {and} \bibinfo{person}{Yiwen Zhang}.} \bibinfo{year}{2018}\natexlab{}.
\newblock \showarticletitle{Imputation of Missing Data in Time Series for Air Pollutants Using Long Short-Term Memory Recurrent Neural Networks}. In \bibinfo{booktitle}{\emph{Proceedings of the 2018 {ACM} International Joint Conference and 2018 International Symposium on Pervasive and Ubiquitous Computing and Wearable Computers, UbiComp/ISWC 2018 Adjunct}}. \bibinfo{publisher}{{ACM}}, \bibinfo{pages}{1293--1300}.
\newblock


\bibitem[Yıldız et~al\mbox{.}(2022)]%
        {9964035}
\bibfield{author}{\bibinfo{person}{A.~Yarkın Yıldız}, \bibinfo{person}{Emirhan Koç}, {and} \bibinfo{person}{Aykut Koç}.} \bibinfo{year}{2022}\natexlab{}.
\newblock \showarticletitle{Multivariate Time Series Imputation With Transformers}.
\newblock \bibinfo{journal}{\emph{IEEE Signal Processing Letters}}  \bibinfo{volume}{29} (\bibinfo{year}{2022}), \bibinfo{pages}{2517--2521}.
\newblock


\bibitem[Zhang and Sennrich(2019)]%
        {DBLP:conf/nips/ZhangS19a}
\bibfield{author}{\bibinfo{person}{Biao Zhang} {and} \bibinfo{person}{Rico Sennrich}.} \bibinfo{year}{2019}\natexlab{}.
\newblock \showarticletitle{Root Mean Square Layer Normalization}. In \bibinfo{booktitle}{\emph{Advances in Neural Information Processing Systems 32: Annual Conference on Neural Information Processing Systems 2019, NeurIPS 2019}}. \bibinfo{pages}{12360--12371}.
\newblock


\bibitem[Zhang et~al\mbox{.}(2023)]%
        {DBLP:conf/cikm/ZhangLY23}
\bibfield{author}{\bibinfo{person}{Kai Zhang}, \bibinfo{person}{Chao Li}, {and} \bibinfo{person}{Qinmin Yang}.} \bibinfo{year}{2023}\natexlab{}.
\newblock \showarticletitle{TriD-MAE: {A} Generic Pre-trained Model for Multivariate Time Series with Missing Values}. In \bibinfo{booktitle}{\emph{Proceedings of the 32nd {ACM} International Conference on Information and Knowledge Management, {CIKM} 2023}}. \bibinfo{publisher}{{ACM}}, \bibinfo{pages}{3164--3173}.
\newblock


\bibitem[Zhang et~al\mbox{.}(2025)]%
        {DBLP:conf/aaai/ZhangSCLGY25}
\bibfield{author}{\bibinfo{person}{Tengxue Zhang}, \bibinfo{person}{Yang Shu}, \bibinfo{person}{Xinyang Chen}, \bibinfo{person}{Yifei Long}, \bibinfo{person}{Chenjuan Guo}, {and} \bibinfo{person}{Bin Yang}.} \bibinfo{year}{2025}\natexlab{}.
\newblock \showarticletitle{Assessing Pre-Trained Models for Transfer Learning Through Distribution of Spectral Components}. In \bibinfo{booktitle}{\emph{AAAI-25}}. \bibinfo{pages}{22560--22568}.
\newblock


\bibitem[Zhang et~al\mbox{.}(2021)]%
        {DBLP:journals/access/ZhangWPQLFW21}
\bibfield{author}{\bibinfo{person}{Zhendong Zhang}, \bibinfo{person}{Chao Wang}, \bibinfo{person}{Xiaosheng Peng}, \bibinfo{person}{Hui Qin}, \bibinfo{person}{Hao Lv}, \bibinfo{person}{Jialong Fu}, {and} \bibinfo{person}{Hongyu Wang}.} \bibinfo{year}{2021}\natexlab{}.
\newblock \showarticletitle{Solar Radiation Intensity Probabilistic Forecasting Based on K-Means Time Series Clustering and Gaussian Process Regression}.
\newblock \bibinfo{journal}{\emph{{IEEE} Access}}  \bibinfo{volume}{9} (\bibinfo{year}{2021}), \bibinfo{pages}{89079--89092}.
\newblock


\bibitem[Zhao et~al\mbox{.}(2023)]%
        {DBLP:journals/pvldb/ZhaoGCHZY23}
\bibfield{author}{\bibinfo{person}{Kai Zhao}, \bibinfo{person}{Chenjuan Guo}, \bibinfo{person}{Yunyao Cheng}, \bibinfo{person}{Peng Han}, \bibinfo{person}{Miao Zhang}, {and} \bibinfo{person}{Bin Yang}.} \bibinfo{year}{2023}\natexlab{}.
\newblock \showarticletitle{Multiple Time Series Forecasting with Dynamic Graph Modeling}.
\newblock \bibinfo{journal}{\emph{Proc. {VLDB} Endow.}} \bibinfo{volume}{17}, \bibinfo{number}{4} (\bibinfo{year}{2023}), \bibinfo{pages}{753--765}.
\newblock


\bibitem[Zhou et~al\mbox{.}(2021)]%
        {DBLP:conf/aaai/ZhouZPZLXZ21}
\bibfield{author}{\bibinfo{person}{Haoyi Zhou}, \bibinfo{person}{Shanghang Zhang}, \bibinfo{person}{Jieqi Peng}, \bibinfo{person}{Shuai Zhang}, \bibinfo{person}{Jianxin Li}, \bibinfo{person}{Hui Xiong}, {and} \bibinfo{person}{Wancai Zhang}.} \bibinfo{year}{2021}\natexlab{}.
\newblock \bibinfo{title}{Informer: Beyond Efficient Transformer for Long Sequence Time-Series Forecasting}.
\newblock \bibinfo{numpages}{11106--11115}~pages.
\newblock


\end{thebibliography}
\clearpage

\section{Appendix}
\label{sec:apdx}
\subsection{Experiment Details}
\subsubsection{Dataset Descriptions}
\label{app:datadetail}
In this part, we give a brief introduction about the datasets in our experiments.

\textbf{Air quality}: The air quality dataset contains PM2.5 data from 36 monitor stations in Beijing, which is sampled hourly for 12 months. There are 13.3\% of missing values with a non-random missing pattern. The air quality dataset contains artificial ground truth with structured missing pattern.

\textbf{Healthcare~(Physionet)}: The healthcare dataset contains 4000 irregularly-sampled clinical time series made up of 35 variables (such as Albumin and heart-rate) for 48 hours collected from ICU. To be consistent with previous studies, the dataset is processed hourly to get 48 timesteps and the processed dataset contains near 80\% missing values without ground truth. For evaluation, we randomly choose 10/50/90\% of the observed values as the ground truth of test dataset.

\textbf{MuJoCo}: The MuJoCo dataset \cite{DBLP:conf/nips/RubanovaCD19} collects a total of 10,000 simulations of the "Hopper" model from the DeepMind Control Suite and MuJoCo simulator. The position of the body in 2D space is uniformly sampled from the interval $[0, 0.5]$. The relative position of the limbs is sampled from the range $[-2, 2]$, and initial velocities are sampled from the interval $[-5, 5]$. In all, there are 10000 sequences of 100 regularly sampled time points with a feature dimension of 14 and a random split of $80/20$ is done for training and testing. We follow the same preprocessing as in \cite{shan2023nrtsi} for fair comparison.

\subsubsection{Hyperparameters}
\label{app:hpara}
Table \ref{tab:hyperp} lists the hyperparameters in SSD-TS.
\begin{table}[h!]
\caption{Hyperparameters in SSD-TS }
\label{tab:hyperp}
\begin{tabular}{l|c}
\hline
                                                                   & SSD-TS \\ \hline
Sequence dim ($C$ in Fig.\ref{fig:modelarch})                      & 128    \\
Residual channels ($K$ in Fig.\ref{fig:modelarch})                 & 128    \\
Num channels (dim of input projections before $\epsilon_{\theta}$) & 128    \\
Diffusion embedding dim                                            & 128    \\
Training iteration                                                 & 150k   \\
Num of conditional SMM blocks                              & 1      \\
Num of input SMM blocks                                    & 1      \\
Num of sequential SMM blocks                               & 1      \\ \hline
\end{tabular}
\end{table}
\subsubsection{Time Series Forecasting Results}
Our model is applicable to time series forecasting tasks and we test it on the ETTm1~\cite{DBLP:conf/aaai/ZhouZPZLXZ21} dataset with forecasting length = 24 and 96, the results are in Table.\ref{tab:tsf}. Our model ranks 1st in 3 of 4 metrics and 2nd on MSE of forecasting length 24, while also performs best among imputation models.

\begin{table}[h!]
\caption{Time series forecasting results on ETTm1 dataset}
\label{tab:tsf}
\resizebox{\linewidth}{!}{
\begin{tabular}{c|cc|cc}
\hline
Forecasting Length & \multicolumn{2}{c|}{24}                   & \multicolumn{2}{c}{96}                    \\ \hline
Model              & MAE                 & MSE                 & MAE                 & MSE                 \\
LSTNet             & 1.170               & 1.968               & 1.542               & 2.762               \\
LSTMa              & 0.629               & 0.621               & 0.913               & 1.339               \\
Reformer           & 0.607               & 0.724               & 0.945               & 1.433               \\
LogTrans           & 0.412               & 0.419               & 0.792               & 0.768               \\
Informer           & 0.369               & \textbf{0.323}           & 0.614               & 0.678               \\
CSDI               & 0.370               & 0.354               & 0.756               & 1.468               \\
Autoformer         & 0.403               & 0.383               & $\underline{0.463}$ & $\underline{0.481}$ \\
SSSD               & $\underline{0.361}$ & 0.351               & 0.547               & 0.538               \\
SSD-TS(Ours)       & \textbf{0.282}           & $\underline{0.331}$ & \textbf{0.391}           & \textbf{0.421}           \\ \hline
\end{tabular}}
\end{table}

\subsubsection{Block Missing Results}
To evaluate the performance of our models on block missing scenario, we trained our models on the PTB-XL dataset (in \cite{DBLP:journals/tmlr/AlcarazS23}) and the result of 20\% block missing is presented in Table.\ref{tab:block}.  

\begin{table}[h!]
\caption{Block missing results on PTB-XL dataset. }
\label{tab:block}
\begin{tabular}{c|cc}
\hline
Model    & MAE                  & RMSE                 \\ \hline
LAMC     & 0.0840               & 0.1171               \\
CSDI     & 0.1054               & 0.2254               \\
DiffWave & 0.0451               & 0.1378               \\
SSSD     & $\underline{0.0324}$ & $\underline{0.0832}$ \\
SSD-TS   & \textbf{0.022}            & \textbf{0.059}            \\ \hline
\end{tabular}
\end{table}

\subsubsection{Additional Time Series Imputation Results}

We also compare our SSD-TS with FIM~\cite{DBLP:conf/iclr/SeifnerCKS25}(on Physionet), ImputeFormer~\cite{DBLP:conf/kdd/NieQMMS24}(on AQI) and Bayotide~\cite{DBLP:conf/icml/FangWLZ024}(on Solar dataset). As the source code of these baselines are not provided, all the results are collected from their corresponding papers. For FIM model, we report its best results.

\begin{table}[h!]
\caption{Comparison with FIM on physionet dataset}
\label{tab:fim}
\begin{tabular}{c|ccccc}
\hline
Models & BRITS & SAITS & CSDI  & FIM-$l(o.n.=4)$ & SSD-TS    \\ \hline
MAE    & 0.297 & 0.257 & \underline{0.252} & 0.402                     & \textbf{0.217} \\ \hline
\end{tabular}
\end{table}
\begin{table}[h!]
\caption{Comparison with Imputeformer on AQI dataset}
\label{tab:impformer}
\resizebox{\linewidth}{!}{
\begin{tabular}{c|cccccccc}
\hline
Models & BRITS & SAITS & TIDER & GRIN  & SPIN  & ImputeFormer & CSDI & SSD-TS   \\ \hline
MAE    & 14.74 & 19.79 & 32.85 & 12.08 & 11.89 & 11.58        & \underline{9.60} & \textbf{6.75} \\ \hline
\end{tabular}}
\end{table}

\begin{table}[h!]
\caption{Comparison with Bayotide on Solar dataset (50\% missing rate)}
\label{tab:bay50}
\resizebox{\linewidth}{!}{
\begin{tabular}{c|cccccccc}
\hline
Models & BRITS & NAOMI & SAITS & TIDER & CSDI  & CSBI  & BayoTIDE & SSD-TS    \\ \hline
RMSE   & 2.842 & 2.918 & 2.791 & \underline{1.679} & 2.276 & 2.097 & 1.699    & \textbf{0.868} \\
MAE    & 1.985 & 2.112 & 1.827 & 0.838 & 0.804 & 1.033 & \underline{0.734}    & \textbf{0.617} \\ \hline
\end{tabular}}
\end{table}

\begin{table}[h!]
\caption{Comparison with Bayotide on Solar dataset (30\% missing rate)}
\label{tab:bay70}
\resizebox{\linewidth}{!}{
\begin{tabular}{c|cccccccc}
\hline
Models & BRITS & NAOMI & SAITS & TIDER & CSDI  & CSBI  & BayoTIDE & SSD-TS    \\ \hline
RMSE   & 2.617 & 2.702 & 2.359 & 1.676 & 2.132 & 1.987 & \underline{1.621}    & \textbf{0.987} \\
MAE    & 1.861 & 2.003 & 1.575 & 0.874 & 1.045 & 0.926 & \underline{0.709}    & \textbf{0.648} \\ \hline
\end{tabular}}
\end{table}
\end{document}